\documentclass[twoside,11pt]{article}

\usepackage{jmlr2e}
\usepackage{times}
\usepackage{url}
\usepackage{caption}
\usepackage{subcaption}
\usepackage[super]{nth}
\usepackage{multirow}
\usepackage{mdframed}
\usepackage{placeins}
\mdfsetup{nobreak=true}
\hypersetup{hidelinks}

\usepackage{amsmath,amsfonts,bm,mathtools,xargs}
\usepackage{scalerel,stackengine}
\PassOptionsToPackage{pdftex}{graphicx}

\def\Figref#1{Figure~\ref{#1}}

\def\Twofigref#1#2{Figures~\ref{#1} and~\ref{#2}}

\def\Secref#1{Section~\ref{#1}}

\def\Foursecrefs#1#2#3#4{Sections~\ref{#1},~\ref{#2},~\ref{#3} and~\ref{#4}}
\def\eqref#1{equation~\ref{#1}}

\def\Eqref#1{Equation~\ref{#1}}

\def\Defref#1{Definition~\ref{#1}}

\def\Twodefrefs#1#2{Definitions~\ref{#1} and~\ref{#2}}

\def\Propref#1{Proposition~\ref{#1}}

\def\Twoproprefs#1#2{Propositions~\ref{#1} and~\ref{#2}}

\def\Thmref#1{Theorem~\ref{#1}}
\def\Twothmrefs#1#2{Theorems~\ref{#1} and~\ref{#2}}

\def\Corolref#1{Corollary~\ref{#1}}

\def\Appref#1{Appendix~\ref{#1}}

\def\Thref#1{Theorem~\ref{#1}}

\def\1{\bm{1}}

\def\rvepsilon{{\boldsymbol{\epsilon}}}

\def\rvmu{{\boldsymbol{\mu}}}
\def\rvsigma{{\boldsymbol{\sigma}}}

\def\rvr{{\mathbf{r}}}

\def\rvx{{\mathbf{x}}}

\def\rvz{{\mathbf{z}}}

\def\vzero{{\bm{0}}}

\def\vmu{{\bm{\mu}}}
\def\vtheta{{\bm{\theta}}}
\def\vphi{{\bm{\phi}}}

\def\mA{{\bm{A}}}

\def\mI{{\bm{I}}}

\def\mS{{\bm{S}}}

\def\mU{{\bm{U}}}
\def\mV{{\bm{V}}}

\def\mX{{\bm{X}}}

\def\mSigma{{\bm{\Sigma}}}

\DeclareMathAlphabet{\mathsfit}{\encodingdefault}{\sfdefault}{m}{sl}
\SetMathAlphabet{\mathsfit}{bold}{\encodingdefault}{\sfdefault}{bx}{n}

\def\sV{{\mathbb{V}}}

\newcommand{\N}{\mathcal{N}}
 \newcommand{\I}{\mathrm{I}}
\newcommand{\E}{\mathbb{E}}

\newcommand{\R}{\mathbb{R}}

\newcommand{\kld}{D_\mathrm{KL}}
\DeclarePairedDelimiterXPP\kl[2]{\kld}(){}{#1\;\delimsize\|\;#2}

\newcommand{\klBig}[2]{D_\mathrm{KL}\Big(#1\;\big\|\;#2\Big)}
\newcommand{\klBigg}[2]{D_\mathrm{KL}\Bigg(#1\;\bigg\|\;#2\Bigg)}
\newcommand{\klbig}[2]{D_\mathrm{KL}\big(#1\;\|\;#2\big)}
\newcommand{\klbigg}[2]{D_\mathrm{KL}\bigg(#1\;\Big\|\;#2\bigg)}

\newcommandx{\Dbig}[4][3,4,usedefault]{D^{#4}_{\mathrm{#3}}\big(#1\;\|\;#2\big)}
\newcommandx{\DBig}[4][3,4,usedefault]{D^{#4}_{\mathrm{#3}}\Big(#1\;\big\|\;#2\Big)}
\newcommandx{\Dbigg}[4][3,4,usedefault]{D^{#4}_{\mathrm{#3}}\bigg(#1\;\Big\|\;#2\bigg)}
\newcommandx{\DBigg}[4][3,4,usedefault]{D^{#4}_{\mathrm{#3}}\Bigg(#1\;\bigg\|\;#2\Bigg)}

\newcommand{\Var}{\mathrm{Var}}

\newcommand{\Cov}{\mathrm{Cov}}

\newcommandx{\ELBO}[2][1,2,usedefault]{\mathcal{L}_{#1}(\vtheta, \vphi; \rvx #2)}

\DeclareMathOperator{\Tr}{Tr}

\makeatletter
\def\biggg{\bBigg@{3.5}}
\def\Biggg{\bBigg@{4}}
\makeatother

\renewcommand\labelenumi{(\roman{enumi})}
\renewcommand\theenumi\labelenumi

\makeatletter
\newcommand{\jmlrdraftheading}{\def\ps@jmlrtps{\let\@mkboth\@gobbletwo%
\def\@oddhead{}\def\@oddfoot{}%
\def\@evenhead{}\def\@evenfoot{}}%
\thispagestyle{jmlrtps}}

\newcommand{\makedrafttitle}{\par
\begingroup
\def\thefootnote{\fnsymbol{footnote}}
\def\@makefnmark{\hbox to 0pt{$^{\@thefnmark}$\hss}}
\@makedrafttitle \@thanks
\endgroup
\setcounter{footnote}{0}
\let\maketitle\relax \let\@makedrafttitle\relax
\gdef\@thanks{}\gdef\@author{}\gdef\@title{}\let\thanks\relax}

\newcommand{\@makedrafttitle}{\vbox{\hsize\textwidth
\linewidth\hsize \vskip \beforetitskip
{\begin{center}
     \Large\bf \@title \par
\end{center}} \vskip \aftertitskip
{\def\and{\unskip\enspace{\rm and}\enspace}%
\def\addr{\small\it}%
\def\email{\hfill\small\sc}%
\def\name{\normalsize\bf}%
\def\AND{\@endauthor\rm\hss \vskip \interauthorskip \@startauthor}
\@startauthor \@author \@endauthor}
\vskip \aftermaketitskip
}}
\makeatother

\jmlrdraftheading

\usepackage{lastpage}
\jmlrdraftheading
\ShortHeadings{Be More Active!}{Bonheme and Grzes}
\firstpageno{1}

\begin{document}

    \title{Be More Active!\\Understanding the Differences Between Mean and Sampled Representations of Variational Autoencoders}

    \author{\name Lisa Bonheme \email lb732@kent.ac.uk\\
    \addr School of Computing\\
    University of Kent\\
    Canterbury, UK
    \AND
    \name Marek Grzes \email m.grzes@kent.ac.uk\\
    \addr School of Computing\\
    University of Kent\\
    Canterbury, UK}

    \editor{David Wipf}

    \makedrafttitle

    \begin{abstract}
        The ability of Variational Autoencoders to learn disentangled representations has made them appealing
        for practical applications.~However, their mean representations, which are generally used for downstream tasks,
        have recently been shown to be more correlated than their sampled counterpart, on which disentanglement
        is usually measured.
        In this paper, we refine this observation through the lens of selective posterior collapse, which states that only
        a subset of the learned representations, the active variables, is encoding useful information while the rest (the passive variables) is discarded.
        We first extend the existing definition to multiple data examples and show that active variables are equally disentangled in mean and sampled
        representations.~Based on this extension and the pre-trained models from \texttt{disentangle\-ment\_lib}, we then isolate the
        passive variables and show that they are responsible for the discrepancies between mean and sampled representations.
        Specifically, passive variables exhibit high correlation scores with other variables in mean representations
        while being fully uncorrelated in sampled ones.
        We thus conclude that despite what their higher correlation might suggest, mean representations are still good
        candidates for downstream tasks applications.~However, it may be beneficial to
        remove their passive variables, especially when used with models sensitive to correlated features.
    \end{abstract}


    \section{Introduction}\label{sec:intro}
    Variational Autoencoders (VAEs) are considered state-of-the-art techniques to learn unsupervised disentangled representations,
    that is, representations encoding separately the different factors of variations~\citep{Bengio2013}.
    ~Disentangled representations are very attractive in terms of interpretability and fairness
    ~\citep{Locatello2019b}, and can be beneficial for downstream tasks such as abstract reasoning~\citep{Steenkiste2019}.\par

    Over the years, multiple regularisation techniques have been developed to encourage
    disentanglement with a specific focus on enforcing the learned latent factors to be uncorrelated.
    As we will discuss in \Secref{sec:background}, while this regularisation is done on the sampled aggregated posterior,
    the learned representation is generally taken to be the mean vector of the posterior distribution.
    However,~\citet{Locatello2019a} reported an increased
    total correlation (TC) and averaged mutual information (MI) over the dimensions of the mean representation compared to
    the results obtained on its sampled counterpart.~This finding raises questions on whether mean representations would
    still benefit from the appealing attributes of disentanglement since sampled representations were shown to be less
    correlated, and thus more disentangled.\par

    Another line of research has shown that VAEs are behaving in a polarised regime, also known as selective posterior collapse
    ~\citep{Dai2018, Rolinek2019}.~In this regime, the relevant dimensions of the sampled representations
    (the active variables) are used by the decoder for reconstruction while the remaining dimensions (the passive variables)
    are `shut down' to closely match the prior.
    However, the polarised regime has only been studied in the context of single data examples for sampled representations.
    Therefore, in \Secref{sec:polarised-regime}, we extend the existing definition of the polarised regime to multiple data examples
    and explore the implications for mean and variance representations. Assuming that VAEs producing disentangled
    representations are behaving in a polarised regime, we show, based on this extended version, that
    active variables of mean representations should not be more correlated than the sampled ones.
    Thus, we argue that the higher correlation reported by~\citet{Locatello2019a} is due to the impact of the passive
    variables on the metrics used.~We verify this hypothesis empirically in~\Secref{sec:truncation-experiment}, and provide
    further analytical justifications in~\Secref{sec:passive-variables}.\par

    Our contribution is three-fold: (1) we extend the definition of the polarised regime to mean and variance representations
    using multiple data examples. (2) we use this extended version to show that the discrepancies between mean and sampled representations observed
    by~\citet{Locatello2019a} are mostly due to the impact of the polarised regime, and especially of the passive
    variables. (3) we explain why passive variables are leading to higher TC and averaged MI scores.
    The code of our experiments is available at~\url{https://github.com/bonheml/tc_study}.

    \paragraph{Notational considerations} Throughout this paper, we use the superscript $^{(i)}$ to denote the values obtained
    for the i$^{th}$ sample $\rvx^{(i)}$ of the random variable $\rvx$, and represent the j$^{th}$ dimension of a vector representation using the subscript
    $_j$. For example, given a random variable $\rvepsilon$ distributed according to $\N(\vzero,\mI)$, $\rvepsilon_j^{(i)}$ is the j$^{th}$ dimension of the
    sample of $\rvepsilon$ obtained for $\rvx^{(i)}$.
    While the mean and diagonal variance representations are functions of $\rvx$ and the network parameters $\phi$, we use a
    shortened version when the meaning is clear from the context, such that $\rvmu \triangleq \rvmu(\rvx;\phi)$, and $\rvsigma \triangleq diag[\mSigma(\rvx;\phi)]$.
    Similarly, for a specific sample $\rvx^{(i)}$, $\rvmu^{(i)} \triangleq \rvmu(\rvx^{(i)};\phi)$, and $\rvsigma^{(i)} \triangleq diag[\mSigma(\rvx^{(i)};\phi)]$.
    We adopt the same notation for the sampled representation, such that $\rvz \triangleq \rvmu + \rvepsilon \rvsigma^{1/2}$, and
    $\rvz^{(i)} \triangleq \rvmu^{(i)} + \rvepsilon^{(i)} (\rvsigma^{(i)})^{1/2}$.

    \section{Background}\label{sec:background}

    \begin{sloppypar}
        \subsection{Variational Autoencoders}
        Variational Autoencoders (VAEs)~\citep{Kingma2013,Rezende2015}
        are deep probabilistic generative models based on variational inference.~The encoder maps some input $\rvx^{(i)}$ to
        a latent representation $\rvz^{(i)}$, and the decoder uses these latent variables to generate an output $\hat{\rvx}^{(i)}$ similar to $\rvx^{(i)}$.
        This can be optimised by maximising the evidence lower bound (ELBO):
        \begin{equation}
            \label{eq:elbo}
            \ELBO = \underbrace{\E_{q_\vphi(\rvz|\rvx)}\big[\log p_\vtheta(\rvx|\rvz)\big]}_{\text{reconstruction term}} -
            \underbrace{ \klbig{q_{\vphi}(\rvz|\rvx)}{p(\rvz)} }_{\text{regularisation term}}.
        \end{equation}
        Generally, $q_\vphi(\rvz|\rvx)$ and $p(\rvz)$ are modelled as multivariate Gaussian distributions to permit closed
        form computation of the regularisation term~\citep{Doersch2016}.
    \end{sloppypar}
    \par
\begin{figure}[h!]
        \centering
        \includegraphics[width=.9\linewidth]{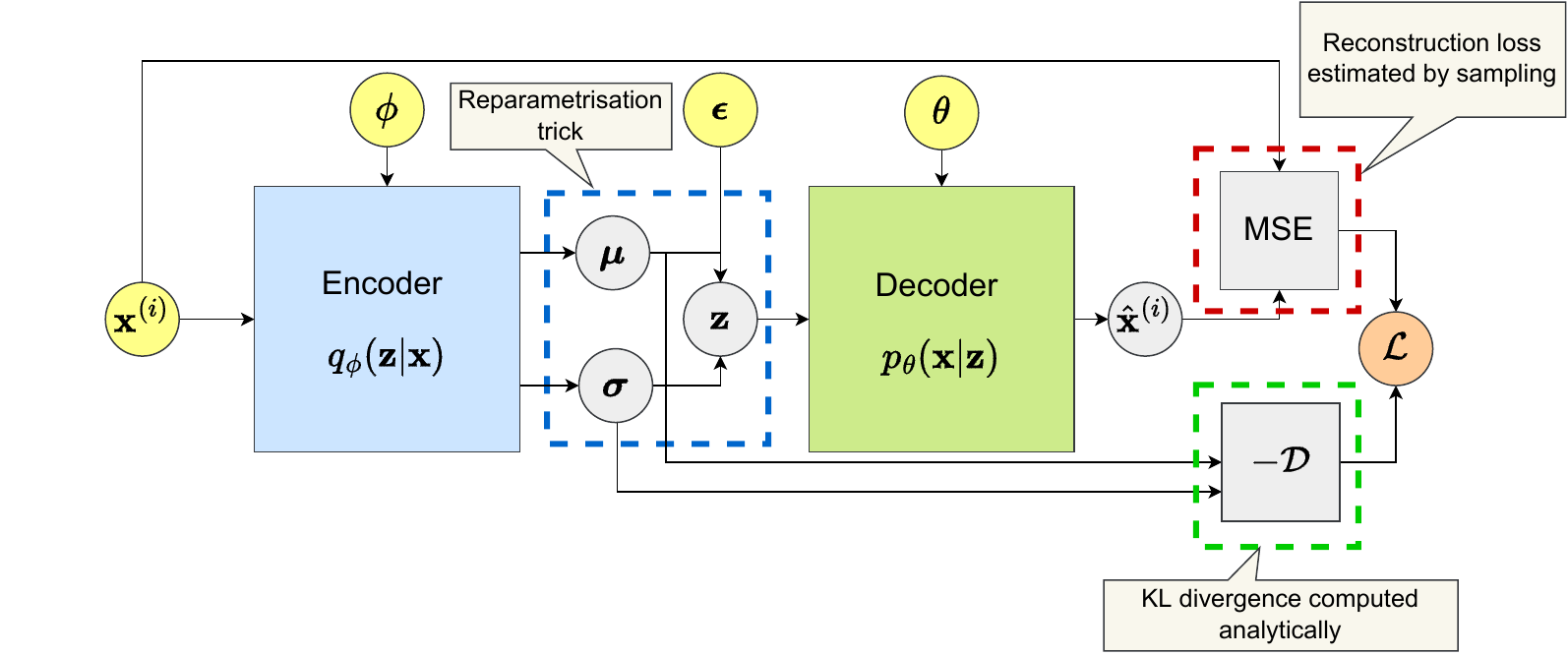}
        \caption{Illustration of a VAE during the training process. The distributions are assumed to be multivariate Gaussian, $\rvmu$
            is the mean layer and $\rvsigma$ is the variance layer. $\rvmu$ and $\rvsigma$ are the parameters of the posterior over $\rvz$.}
        \label{fig:vae-archi}
    \end{figure}

    As illustrated in~\Figref{fig:vae-archi}, given the mean $\rvmu$ and diagonal covariance $\rvsigma$ of a random variable $\rvx$,
    the sampled representation is obtained using the reparameterisation trick~\citep{Kingma2013} such that
    $\rvz = \rvmu + \rvsigma^{1/2}\rvepsilon$ where $\rvepsilon$ is a random variable with a Gaussian distribution $\N(\vzero,\mI)$.\par
\begin{figure}[h!]
        \centering
        \includegraphics[ width=.38\linewidth]{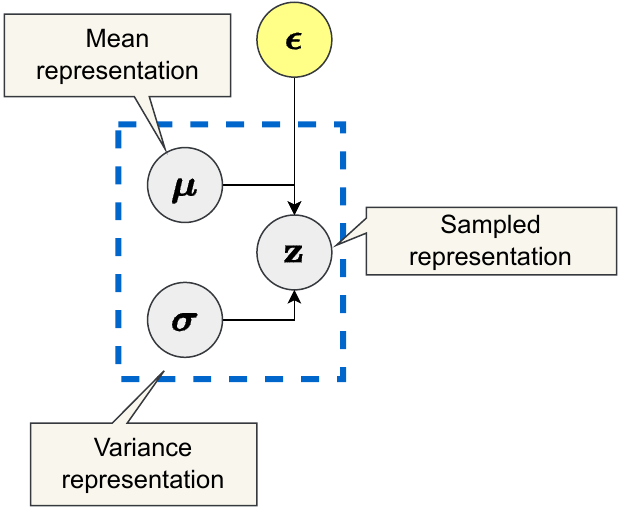}
        \caption{Mean, variance, and sampled representations of VAEs.}
        \label{fig:vae-representations}
    \end{figure}

    In this paper we are interested in investigating the discrepancies between
    the mean and sampled representations, that is $\rvmu$ and $\rvz$ respectively.
    ~Specifically, our goal is to explain the higher correlation of mean representations reported by~\citet{Locatello2019a}.
    In the following sections, the representations learned by the mean layer will be
    referred to as mean representations, and those learned during the sampling stage as sampled representations,
    as per~\Figref{fig:vae-representations}.

    \subsection{Disentangled representation learning with VAEs}\label{subsec:disentangled-vaes}
    While various learning objectives and architectures have been proposed for VAEs, we will focus on the family of
    methods increasing the weight on the regularisation term of~\Eqref{eq:elbo} to produce disentangled representations,
    similarly to~\citet{Locatello2019a}.~We refer the reader to~\citet{Tschannen2018} for a broader overview of the
    existing VAE architectures.\par

    \paragraph{$\beta$-VAE}\citet{Higgins2017} introduced a new learning objective whose goal was to bias the
    encoding-reconstruction trade-off by penalising the regularisation term more strongly.
    This is formulated as the following learning objective:
    \begin{equation}
        \label{eq:Higgins2017}
        \E_{q_{\vphi}(\rvz|\rvx)}\big[\log p_{\vtheta}(\rvx|\rvz)\big]-\beta\klbig{q_{\vphi}(\rvz|\rvx)}{p_{\vtheta}(\rvz)}.
    \end{equation}
    \Eqref{eq:Higgins2017} is similar to the original VAE objective seen in \Eqref{eq:elbo} with the addition of the $\beta$ parameter which,
    when $\beta>1$, increases the bias on the encoding optimisation.
    One of the downsides of penalising the encoding more strongly is that the reconstruction is of lower quality.

    \begin{sloppypar}
        \paragraph{Annealed VAE}\citet{Burgess2018} provided an analysis of $\beta$-VAE disentangled representations through the lens of information
        theory, based on the learning objective described by~\citet{Alemi2017}.
        They argue that because $\beta$-VAE is increasing the pressure on the encoding capacity of the network, the optimal way
        to encode information would be on separate dimensions, leading to disentanglement.
        They hypothesise that $\beta$-VAE will learn the latent variables having the most impact on the reconstruction first, then
        gradually optimise less critical variables.
        To ease the learning of these less important latent variables, they propose to gradually increase the encoding capacity
        during the training process, relaxing the initial constraint.
        This leads to the following objective, where \textrm{C} is a parameter that can be understood as a channel capacity and $\gamma$ is a hyper-parameter
        penalising the divergence, similarly to $\beta$ in $\beta$-VAE:
        \begin{equation}
            \E_{q_{\vphi}(\rvz|\rvx)}\big[\log p_{\vtheta}(\rvx|\rvz)\big]-\gamma\left|\klbig{q_{\vphi}(\rvz|\rvx)}{p_{\vtheta}(\rvz)}-\mathrm{C}\right|\label{eq:Burgess2018}.
        \end{equation}
        As the training progresses, the channel capacity $\mathrm{C}$ is increased, going from zero to its maximum channel capacity $\mathrm{C_{max}}$.
        For example, given a maximum channel capacity of 100, during the first training step, any deviation from the KL divergence will be penalised similarly to $\beta$-VAE because $\mathrm{C} = 0$.
        After $n$ steps, once the channel capacity will be annealed to its maximum value, the KL divergence will be penalised only when it is higher than 100.
        VAEs that use~\Eqref{eq:Burgess2018} as a learning objective will be referred to as annealed VAEs in the rest of this paper.
    \end{sloppypar}

    \paragraph{Factor VAE} In Factor VAE,~\citet{Kim2018} addressed disentanglement learning by enforcing the independence of the latent factors
    learned by VAEs. This approach is slightly different from the $\beta$-VAE objective presented in
    ~\Eqref{eq:Higgins2017} as it takes into account the KL divergence $\klbig{q_{\vphi}(\rvz)}{p_{\vtheta}(\rvz)}$ to further
    decompose the expected value over the data distribution of the second term of the loss function as:
    \begin{equation*}
        \E_{p_{\vtheta}(\rvx)}\big[\klbig{q_{\vphi}(\rvz|\rvx)}{p_{\vtheta}(\rvz)}\big]=\I_{e}(\mathrm{\rvx;\rvz})+\klbig{q_{\vphi}(\rvz)}{p_{\vtheta}(\rvz)}.
    \end{equation*}

    \citet{Kim2018} suggested that $\beta$-VAE's lower data generation quality is due to the penalisation of $\I_{e}(\mathrm{\rvx;\rvz})$,
    the mutual information between the observed and latent variables.
    They argued that only the distance between the estimated latent factors and the prior should be penalised to encourage
    disentanglement, and they proposed a new objective to this end:
    \begin{equation}
        \label{eq:kl-pz-qz}
        \E_{p(\rvx)}\left[\E_{q_{\vphi}(\rvz|\rvx)}\left[\log p_{\vtheta}\big(\rvx|\rvz\big)\right]-\klbig{q_{\vphi}\big(\rvz|\rvx\big)}{p_{\vtheta}(\rvz)}\right]
        -\gamma\klbig{q_{\vphi}(\rvz)}{p_{\vtheta}(\rvz)}.
    \end{equation}
    Here, $\klbig{q_{\vphi}(\rvz)}{p_{\vtheta}(\rvz)}$ is approximated by penalising the dependencies between the dimensions of $q_{\vphi}(\rvz)$:
    \begin{equation}
        \frac{1}{n}\sum_{i=1}^{n}\left[\E_{q_{\vphi}(\rvz|\rvx)}\left[\log p_{\vtheta}\big(\rvx^{(i)}|\rvz\big)\right]-\klbig{q_{\vphi}\big(\rvz|\rvx^{(i)}\big)}{p_{\vtheta}(\rvz)}\right]
        -\underbrace{\gamma\klBigg{q_{\vphi}(\rvz)}{\prod_{j=1}^{\mathrm{d}}q_{\vphi}(\rvz_{j})}}_{\textrm{total correlation}}\label{eq:Kim2018}.
    \end{equation}

    As the computation of the total correlation defined in~\Eqref{eq:Kim2018} is intractable, it is estimated by sampling a batch from $q_{\vphi}(\rvz)$
    and shuffling the values of each dimension of the latent variables to obtain the samples for $\prod_{j=1}^{\mathrm{d}}q_{\vphi}(\rvz_{j})$.
    A binary classifier is then trained to recognise the samples belonging to $q_{\vphi}(\rvz)$, and
    the density ratio is computed using the probability $p_{classif}(\rvz)$ given by the classifier that
    the samples belong to $q_{\vphi}(\rvz)$:
    \begin{equation*}
        \klBigg{q_{\vphi}(\rvz)}{\prod_{j=1}^{\mathrm{d}}q_{\vphi}(\rvz_{j})}\approx\E_{q_{\vphi}(\rvz)}\left[\log\frac{p_{classif}(\rvz)}{1-p_{classif}(\rvz)}\right].
    \end{equation*}
    In~\citet[Appendix F and I]{Kim2018} the authors compared the results obtained with~\Eqref{eq:Kim2018} and~\Eqref{eq:kl-pz-qz} (i.e., where $p_\theta(\rvz)$ is not approximated by
    $\prod_{j=1}^{\mathrm{d}}q_{\vphi}(\rvz_{j})$), but obtained lower disentanglement with the latter.
    They concluded that enforcing a factorised $q_\phi(\rvz)$ was more beneficial for disentanglement than enforcing $q_\phi(\rvz)$ to be as close as possible to $\N(\vzero,\mI)$.

    \paragraph{$\beta$-TC VAE} Similarly to~\citet{Kim2018},~\citet{Chen2018} proposed to optimise~\Eqref{eq:Kim2018}.
    The main difference is that \citet{Chen2018} are approximating the total correlation using mini-batch weighted sampling.
    Here, the estimation is computed over a mini-batch of samples $\{\rvx^{(i)}\}_{i=1}^{m}$ as follows:
    \begin{equation}
        \label{eq:beta-tc-vae}
        \E_{q_{\vphi}(\rvz)}[\log{q_{\vphi}(\rvz)}]\approx\frac{1}{m}\sum_{i=1}^{m}\left(\log \frac{1}{nm}\sum_{k=1}^{m}q_{\vphi}(\rvz^{(i)}|\rvx^{(k)})\right),
    \end{equation}
    where $m$ is the number of samples in the mini-batch, and $n$ total number of input examples.
    $\E_{q_{\vphi}(\rvz_j)}[\log{q_{\vphi}(\rvz_j)}]$ can be computed in a similar way.
    We refer the reader to~\citet[Appendix C.1]{Chen2018} for the detailed derivation of~\Eqref{eq:beta-tc-vae}.

    \paragraph{DIP-VAE} Similarly to~\citet{Kim2018} and~\citet{Chen2018},~\citet{Kumar2018} proposed to regularise the
    distance between $q_\phi(\rvz)$ and $p(\rvz)$ using~\Eqref{eq:kl-pz-qz}.
    The main difference is that here $\klbig{q_{\phi}(\rvz)}{p(\rvz)}$ is measured by matching the moments of the learned
    distribution $q_{\phi}(\rvz)$ and its prior $p(\rvz)$.
    The second moment of the learned distribution is given by:
    \begin{equation}
        \label{eq:Kumar2018Cov}
        \Cov_{q_{\phi}(\rvz)}[\rvz] = \Cov_{p(\rvx)}\left[\rvmu\right] + \E_{p(\rvx)}\left[\rvsigma\right].
    \end{equation}
    Two divergences are then defined.~The first, DIP-VAE I, penalises only the first term of~\Eqref{eq:Kumar2018Cov}:
    \begin{equation*}
        \lambda\klbig{q_{\phi}(\rvz)}{p_{\theta}(\rvz)} = \lambda_{od} \sum_{i\neq j}\left(\Cov_{p(\rvx)}\left[ \rvmu \right]\right)_{ij}^{2}
        + \lambda_{d}\sum_{i} \left( \Cov_{p(\rvx)} \left[ \rvmu \right]_{ii} -1 \right)^{2},
    \end{equation*}
    where $\lambda_{od}$ and $\lambda_{d}$ are the off-diagonal and diagonal regularisation terms, respectively.~The second, DIP-VAE II,
    penalises both terms of~\Eqref{eq:Kumar2018Cov}:
    \begin{equation*}
        \lambda\klbig{q_{\phi}(\rvz)}{p_{\theta}(\rvz)} = \lambda_{od} \sum_{i\neq j}\left(\Cov_{q_{\phi}(\rvz)}\left[\rvz\right]\right)_{ij}^{2}
        + \lambda_{d} \sum_{i} \left(\Cov_{q_{\phi}(\rvz)}\left[\rvz\right]_{ii} -1\right)^{2}.
    \end{equation*}

    ~Note that because DIP-VAE I directly encourages diagonal covariance matrices in the mean representation,
    it will have a low correlation in the mean representation, which, as observed by \citet{Locatello2019a}, mirrors the correlations in the sampled representation.
    Moreover, the discrepancies between mean and sampled representations were observed by \citet{Locatello2019a} in the
    context of methods which explicitly regularise the disentanglement of $\rvz$, but not $\rvmu$.
    In this study, we will thus consider DIP-VAE II, which enforces the covariance matrix of the
    sampled representation to be diagonal, but not DIP-VAE I as it explicitly regularises $\rvmu$.

    \subsection{Benefits of disentanglement on downstream tasks}
    Disentangled representations have been shown to reduce the sample complexity of abstract reasoning tasks~\citep{Steenkiste2019},
    improve the fairness of downstream task models~\citep{Locatello2019b,Creager2019} and their interpretability~\citep{Higgins2017,Adel2018}.
    However, when the mean representations are more correlated than the sampled representations on which the disentanglement is measured,
    it will hamper the interpretability of downstream task models~\citep{Alin2010,Chan2022}, and reduce their fairness~\citep{Locatello2019b,Trauble2021}.
    Moreover, under a certain level of supervision, VAEs can provably provide identifiable representations~\citep{Khemakhem2020,Mita2021}, and it is conjectured that under
    specific constraints, this is also possible in the unsupervised setting~\citep{Reizinger2022}.
    It is thus important to investigate the origin of the discrepancies between mean and sampled representations to determine if one can
    still benefit from disentanglement when using mean representations on downstream tasks.

    \subsection{Related work} The discrepancy between the total correlation of mean and sampled representations
    observed by~\citet{Locatello2019a} have recently been investigated by \citet{Cheng2021} who provide a theoretical
    justification of the higher total correlation scores of the mean representations.
    However, they did not consider the polarised regime which is a necessary
    condition for VAE to provide good reconstruction~\citep{Dai2018, Rolinek2019, Dai2020}.
    Thus, their work is complementary to ours in the case where VAEs are not learning in a polarised regime.

    \section{The polarised regime}\label{sec:polarised-regime}
    The polarised regime, also known as selective posterior collapse, is the ability of VAEs to `shut down' superfluous
    dimensions of their sampled latent representations while providing a high precision on the remaining ones~\citep{Rolinek2019, Dai2020}.
    As a result, the sampled representation can be separated into two subsets of variables, active and passive.
    The active variables correspond to the subset of the sampled latent representation that is needed for the reconstruction.
    They have a low variance, and are close to the mean variables.
    The passive variables correspond to the superfluous dimensions that are discarded by the VAE. They follow a zero-mean unit-variance Gaussian distribution to optimally match the prior and are ignored by the decoder, which only uses the variables that help to reconstruct the input.\par
    The existence of active and passive variables has been shown to be a necessary condition for the VAEs to provide a good reconstruction ~\citep{Dai2018, Dai2020}.
    However, when the weight on the regularisation term of the ELBO given in~\Eqref{eq:elbo} increases, VAEs are pruning more active variables to
    minimise the regularisation loss.~When this weight becomes too large, the representations collapse to the prior,
    containing only passive variables~\citep{Lucas2019,Dai2020}.

    \subsection{The polarised regime for one data example}\label{sec:polarised-regime-xi}

    Based on the definition of \citet{Rolinek2019}, we can characterise the active and passive variables of sampled representations as follows.
    \begin{definition}[Polarised regime]
        \label{def:polarised-regime}
        When a VAE learns in a polarised regime, for a given data example $\rvx^{(i)} \in \mX$, its mean, variance, and sampled representations, $\rvmu^{(i)} \triangleq \vmu(\rvx^{(i)}; \phi)$, $\rvsigma^{(i)} \triangleq diag[\mSigma(\rvx^{(i)}; \phi)]$, and $\rvz^{(i)} \triangleq \rvmu^{(i)} + \rvepsilon^{(i)} (\rvsigma^{(i)})^{1/2}$, respectively, are composed of a set of passive and active variables, $\sV_p^{(i)} \cup \sV_a^{(i)}$ such that, for each data example $\rvx^{(i)}$:
        \begin{enumerate}
            \item $|\vmu_j^{(i)}| \ll 1 \text{, } \rvsigma_j^{(i)} \approx 1 \text{, and } \rvz_j^{(i)} \approx \rvepsilon_j^{(i)} \quad \forall \; j \in \sV_p^{(i)}$,
            \item $\rvsigma_j^{(i)} \ll 1  \text{ and } \rvz_j^{(i)} \approx \rvmu_j^{(i)} \quad \forall \; j \in \sV_a^{(i)}$,
        \end{enumerate}
        where $\rvepsilon^{(i)} \sim \N(\vzero,\mI)$, $j$ is the j$^{\text{th}}$ variable of a representation, and $|\cdot|$ denotes the absolute value.
    \end{definition}

    The polarised regime can also be seen as a sparsity-inducing mechanism which will prune the superfluous columns of the weights of the first layer
    of the decoder~\citep{Dai2017,Dai2018b}. As the corresponding dimensions of the mean and variance representations will not have any influence on the decoder (i.e., they are passive),
    they will only be optimised with respect to the KL divergence, thus becoming close to 0 and 1, respectively.

    While~\Defref{def:polarised-regime} provides an overview of the polarised regime for each data example, it is not readily usable
    to analyse discrepancies between mean and sampled representations as they are observed over the whole dataset. We will thus
    extend the definition of the polarised regime to multiple data examples in~\Secref{sec:polarised-regime-x} and show that the
    discrepancies between mean and sampled representations can only originate from variables which are not active.

    \subsection{Generalisation of the polarised regime to multiple data examples}\label{sec:polarised-regime-x}

\begin{figure}[ht!]
        \centering
        \includegraphics[width=0.5\linewidth]{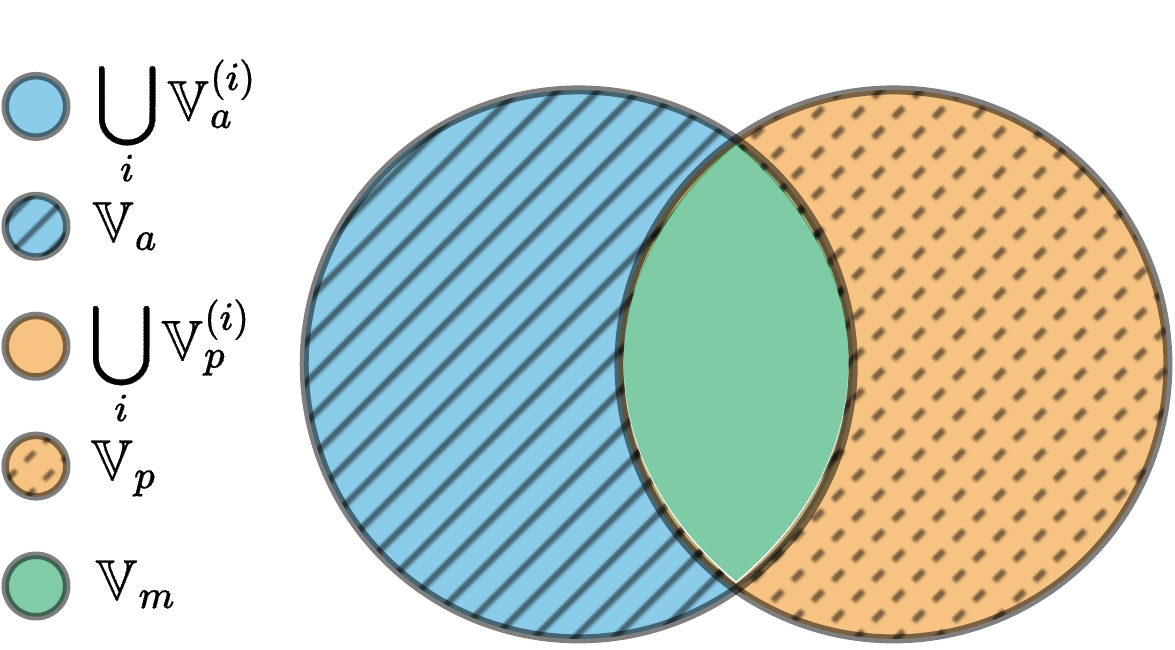}
        \caption{A graphical representation of~\Defref{def:var-type}.}
        \label{fig:var-type}
    \end{figure}

    Now we will propose a new generalisation of~\Defref{def:polarised-regime} to multiple data examples, which will serve as a basis for our analysis in~\Secref{sec:truncation-experiment}.
    Given that a variable can either be active or passive for a given data example, when considering multiple data examples, three cases arise:
    \begin{itemize}
        \setlength\itemsep{-0.1em}
        \item A variable is passive for all the data examples.
        \item A variable is active for all the data examples.
        \item A variable is active for some data examples and passive otherwise.
    \end{itemize}
    These three types of variables are formalised in~\Defref{def:var-type} and illustrated in~\Figref{fig:var-type}.

    \begin{definition}[Variable types]
        \label{def:var-type}
        When considered over multiple data examples $\mX = \{\rvx^{(i)}\}_{i=1}^{n}$, the latent representations are composed of a set of passive, active, and mixed variables $\sV_p \cup \sV_a \cup \sV_m$, which are defined as follows:
        \begin{enumerate}
            \item $\sV_p \triangleq \bigcap\limits_{i=1}^{n}\sV_p^{(i)}$,
            \item $\sV_a \triangleq \bigcap\limits_{i=1}^{n} \sV_a^{(i)}$,
            \item $\sV_m \triangleq \Big(\bigcup\limits_{i=1}^{n} \sV_p^{(i)}\Big) \bigcap \Big(\bigcup\limits_{i=1}^{n} \sV_a^{(i)}\Big)$.
        \end{enumerate}
    \end{definition}

    Based on~\Twodefrefs{def:polarised-regime}{def:var-type}, we will consider the properties of the mean and variance representations over multiple data examples in~\Twoproprefs{prop:polarised-mean}{prop:polarised-var}, then describe their implications for the sampled representations in~\Thmref{thm:polarised-sampled}.

    \begin{proposition}[Polarised regime of $\rvmu$ over $\mX$]
        \label{prop:polarised-mean}
        When a VAE learns in a polarised regime, its mean representation $\rvmu \triangleq \rvmu(\rvx; \phi) \approx \rvmu(\mX; \phi)$ is composed of a set of passive, active and mixed variables $\sV_p \cup \sV_a \cup \sV_m$ such that, over $\mX$:
        \begin{enumerate}
            \item $|\bar{\rvmu}_j| \ll 1 \text{ and } Var(\rvmu_j) \ll 1 \quad \forall \; j \in \sV_p$,
            \item $Var(\rvmu_j) > Var(\rvmu_k)\quad \forall \; j \in \sV_a \;, \forall \; k \in \sV_p$,
        \end{enumerate}
        where $\bar{\rvmu}_j \triangleq \E_{p(\rvx)}[\rvmu_j]$, and $Var(\cdot)$ denotes the variance.
    \end{proposition}

    The first point of~\Propref{prop:polarised-mean} indicates that passive variables of mean representations are almost constant when
    considered over multiple data examples. Indeed, because they are passive for each data example, according to (i) of ~\Defref{def:polarised-regime} they will consistently take values close to 0. Thus, they will have a variance and expected value close to 0.

    Because active variables of the mean representation encode some information about the input, their value will vary depending on the input and thus have a higher variance than their passive counterpart, as stated in (ii) of~\Propref{prop:polarised-mean}.

    Now, let us analyse the effect of the polarised regime on sampled representations over multiple data examples.

    \begin{proposition}[Polarised regime of $\rvsigma$ over $\mX$]
        \label{prop:polarised-var}
        When a VAE learns in a polarised regime, its variance representation $\rvsigma \triangleq diag[\mSigma(\rvx; \phi)] \approx diag[\mSigma(\mX; \phi)]$ is composed of a set of passive, active and mixed variables $\sV_p \cup \sV_a \cup \sV_m$ such that, over $\mX$:
        \begin{enumerate}
            \item $\bar{\rvsigma}_j \approx 1 \text{ and } Var(\rvsigma_j) \ll 1 \quad \forall j \in \sV_p$,
            \item $\bar{\rvsigma}_j \ll 1 \text{ and } Var(\rvsigma_j) \ll 1 \quad \forall j \in \sV_a$,
            \item $Var(\rvsigma_j) < Var(\rvsigma_k) \quad \forall j \notin \sV_m \;, \forall k \in \sV_m $,
        \end{enumerate}
        where $\bar{\rvsigma}_j \triangleq \E_{p(\rvx)}[\rvsigma_j]$.
    \end{proposition}

    We know from~\Defref{def:polarised-regime} that the variance representation is always close to 1 when the variables are passive and to 0 when they are active.
    Thus, variables that are passive (resp. active) over the whole dataset will be almost constant with an expected value close to 1 (resp. 0), as stated in (i) and (ii) of~\Propref{prop:polarised-var}.

    The variance representations of mixed variables will alternate between 1 and 0, depending on whether they are passive or active for the considered data examples.
    Thus, as described in (iii) of ~\Propref{prop:polarised-var}, they will vary more than active and passive variables.

    \begin{theorem}[Polarised regime of $\rvz$ over $\mX$]
        \label{thm:polarised-sampled}
        When a VAE learns in a polarised regime, its sampled representation $\rvz \triangleq \rvmu + \rvepsilon \rvsigma^{1/2}$ is composed of a set of passive, active and mixed variables $\sV_p \cup \sV_a \cup \sV_m$ such that, over $\mX$:
        \begin{enumerate}
            \item $p(\rvz_j) \approx p(\rvepsilon_j) \quad \forall \; j \in \sV_p$,
            \item $p(\rvz_j) \approx p(\rvmu_j)\quad \forall \; j \in \sV_a$,
            \item $p(\rvz_j) = c\; p(\rvepsilon_j) + (1 - c) \; p(\rvmu_j)\quad \forall \; j \in \sV_m$,
        \end{enumerate}
        where $0 < c < 1$.
    \end{theorem}
    The proof can be found in~\Appref{subsec:proof-polarised-sampled}.\\
    Given that active variables are the only type of variables with approximately the same distributions in mean and sampled representations,~\Corolref{cor:polarised-diff} immediately follows.

    \begin{corollary}
        ~\label{cor:polarised-diff}
        Any discrepancies between the mean and sampled representations can only come from the mixed and passive variables.
    \end{corollary}

    \subsection{Empirical demonstration of polarised regimes}\label{subsec:empirical-demo}

\begin{figure}[h!]
        \centering
        \begin{subfigure}{.33\textwidth}
            \centering
            \includegraphics[width=\linewidth]{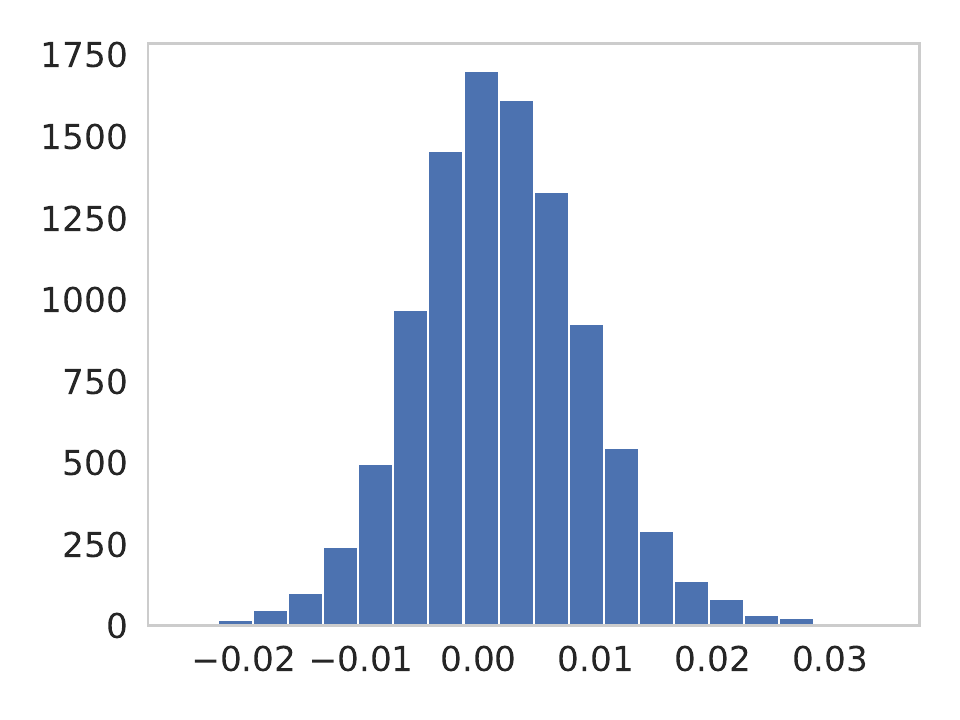}
            \caption{Passive variable}
            \label{fig:passive-var-mean}
        \end{subfigure}\begin{subfigure}{.33\textwidth}
                           \centering
                           \includegraphics[width=\linewidth]{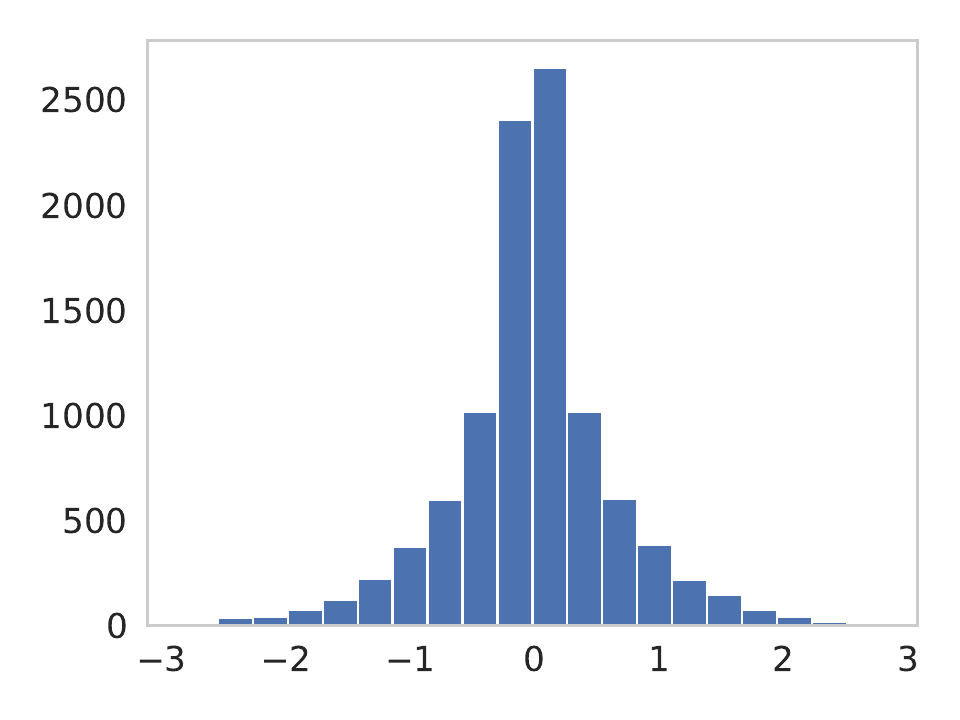}
                           \caption{Mixed variable}
                           \label{fig:mixed-var-mean}
        \end{subfigure}\begin{subfigure}{.33\textwidth}
                           \centering
                           \includegraphics[width=\linewidth]{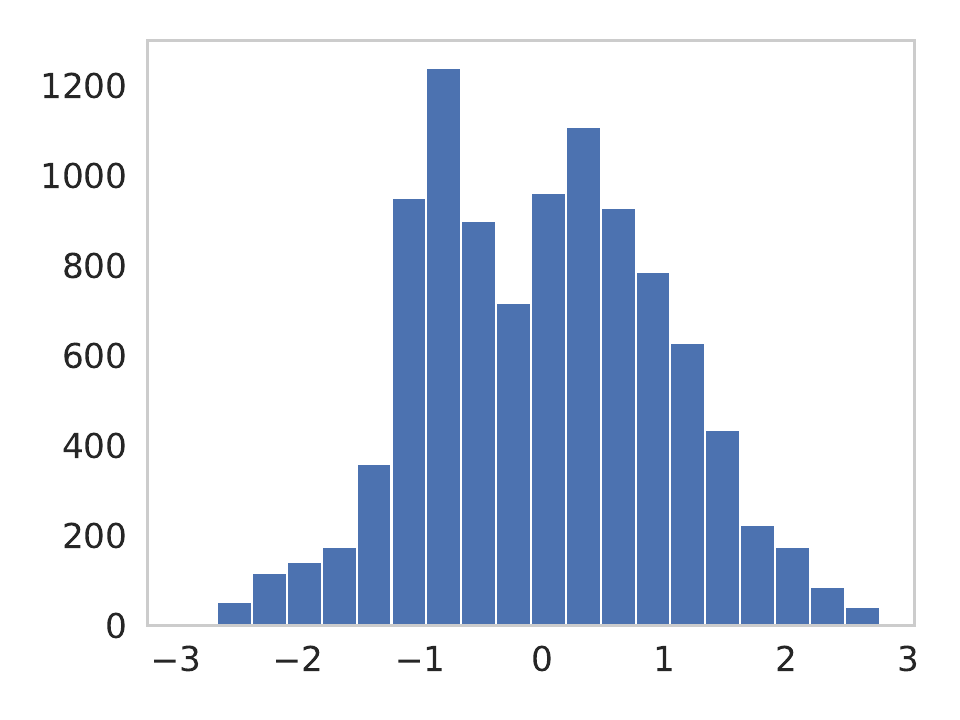}
                           \caption{Active variable}
                           \label{fig:active-var-mean}
        \end{subfigure}
        \caption{Empirical distributions of the \nth{2}, \nth{4}, and \nth{10} latent dimensions of the mean representations
        of a $\beta$-VAE trained on dSprites with $\beta = 8$, which illustrate the typical appearance of passive, mixed, and active variables.
        The histograms are computed using 10000 input examples.}
        \label{fig:var-mean}
    \end{figure}

\begin{figure}[h!]
        \centering
        \begin{subfigure}{.33\textwidth}
            \centering
            \includegraphics[width=\linewidth]{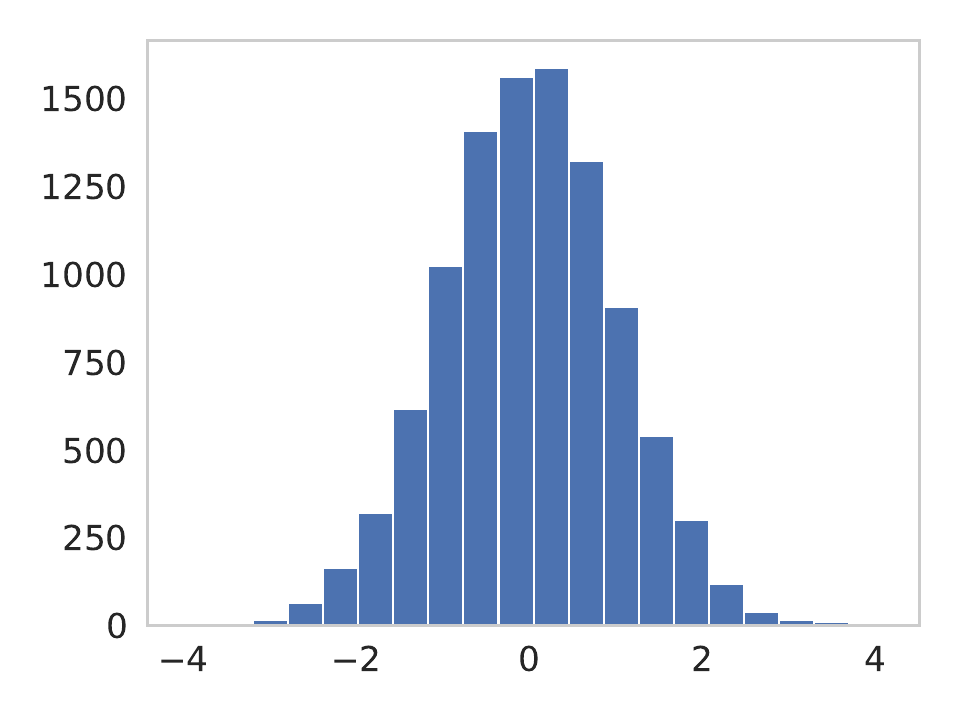}
            \caption{Passive variable}
            \label{fig:passive-var-sampled}
        \end{subfigure}\begin{subfigure}{.33\textwidth}
                           \centering
                           \includegraphics[width=\linewidth]{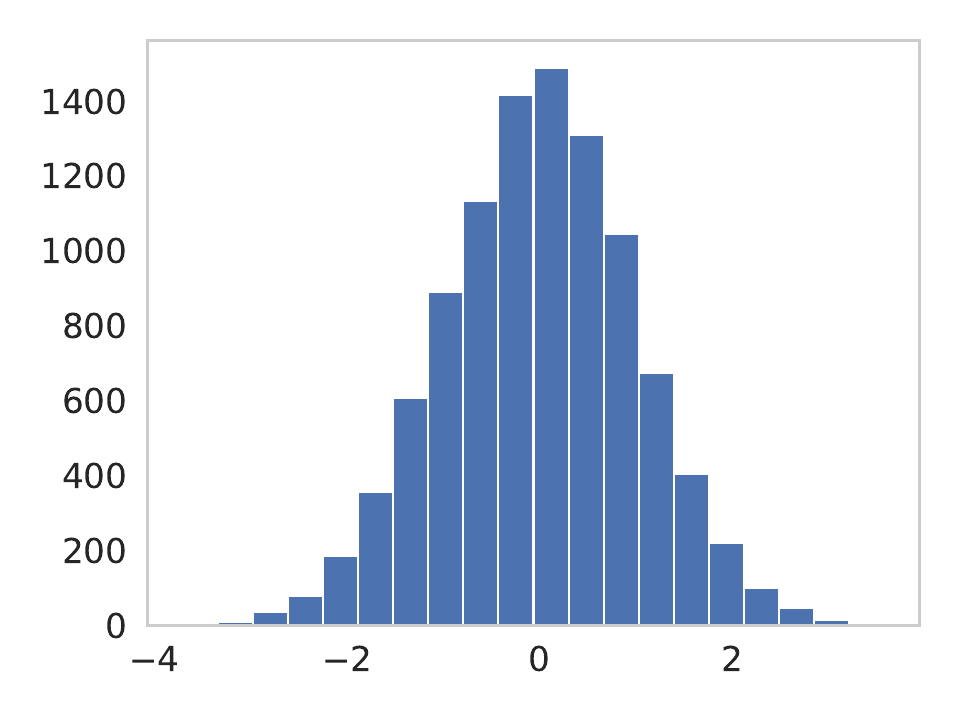}
                           \caption{Mixed variable}
                           \label{fig:mixed-var-sampled}
        \end{subfigure}\begin{subfigure}{.33\textwidth}
                           \centering
                           \includegraphics[width=\linewidth]{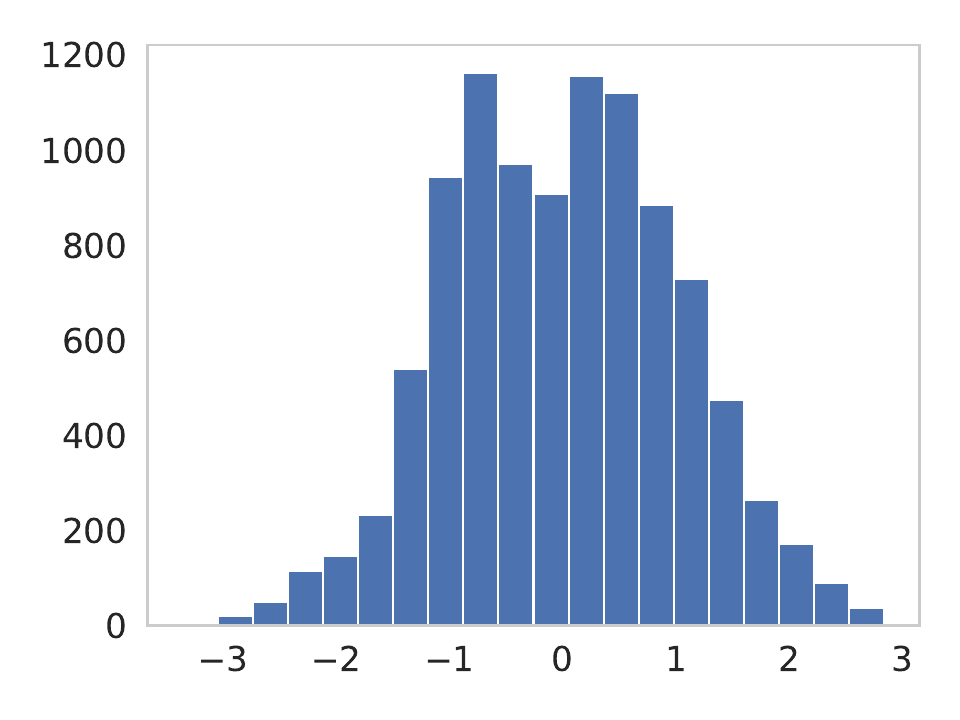}
                           \caption{Active variable}
                           \label{fig:active-var-sampled}
        \end{subfigure}\caption{Empirical distributions of the \nth{2}, \nth{4}, and \nth{10} latent dimensions of the sampled representations
        of a $\beta$-VAE trained on dSprites with $\beta = 8$, which illustrate the typical appearance of passive, mixed, and active variables.
        The histograms are computed using 10000 input examples.}
        \label{fig:var-sampled}
    \end{figure}

    We will now verify that~\Thref{thm:polarised-sampled} holds empirically using a
    $\beta$-VAE trained on the dSprites dataset~\citep{Higgins2017}.
    By comparing the passive variable distribution of mean and sampled representations provided in~\Twofigref{fig:passive-var-mean}{fig:passive-var-sampled},
    we can see that both have a mean of zero, and that the variance of the variable is close to zero in the mean representation, and
    to one in the sampled representation, consistently with statement (i) of~\Thref{thm:polarised-sampled} and~\Propref{prop:polarised-mean}.
    As described in statement (ii) of~\Thmref{thm:polarised-sampled}, the active variable of the mean representation observed in~\Figref{fig:active-var-mean}
    follows a similar distribution as its sampled counterpart in~\Figref{fig:active-var-sampled}.
    \Twofigref{fig:mixed-var-mean}{fig:mixed-var-sampled} show that mixed variables can also be identified. Their
    variance in the mean representation is larger than for passive variables and increases in the sampled representation.
    Moreover, we can see a sharp peak around zero in the mean representation, which is smoothed out in the sampled one.
    ~This likely corresponds to the passive component of the mixed variables, which is close to zero with very low variance in the mean
    representation and to $\N(\vzero,\mI)$ in the sampled representation. Overall, these observations are consistent with a mixture distribution,
    as per statement (iii) of~\Thmref{thm:polarised-sampled}.\par
    ~Note that while the distribution of the passive variables of sampled representations is encouraged to be Gaussian by the KL
    divergence term of the ELBO, we can see in~\Twofigref{fig:var-mean}{fig:var-sampled} that the mixed and active variables are not guaranteed
    to be Gaussian in mean and sampled representations.
    Indeed, as they convey some information about the data, their distribution can be arbitrarily different from the prior
    and may or may not be Gaussian.~The resulting increased KL divergence of active variables is then compensated by passive
    variables that match the prior exactly.
    Moreover, the distribution of the passive variables of mean representations may also not be Gaussian as it is only optimised
    to have low variance.

    These observations also provide empirical evidence of~\Corolref{cor:polarised-diff}, that only the active variables are similar
    in mean and sampled representations.~When the regularisation strength increases, the sampled and mean representations
    will contain more passive and mixed variables, hence \textit{the higher the regularisation weight, the greater the
    difference between the mean and sampled representations}.~This is the main insight that will allow us to explain the
    discrepancies between the correlation scores of the mean and sampled representations observed by \citet{Locatello2019a}.

    \section{Assessing the impact of mixed and passive variables on the differences between mean and sampled representations}\label{sec:truncation-experiment}
    In~\Thref{thm:polarised-sampled} and~\Secref{sec:polarised-regime} we have seen that while active variables are equivalent in mean and sampled representations,
    it is not the case for mixed and passive variables.~As the number of non-active variables increases with the regularisation
    strength, we hypothesise that they may be the source of the stronger correlation of mean representations observed by
    \citet{Locatello2019a}.\FloatBarrier\par
    To verify this hypothesis, we investigate if active variables alone are equivalently correlated in mean and sampled
    representations and how passive variables would impact the different metrics used.~To do so, we divide our experiment into four steps:
    \begin{enumerate}
        \item Using~\Propref{prop:polarised-var}, identify the type of variable (active, mixed or passive) stored at each index of the representations considered.
        \item Based on~\Propref{prop:polarised-mean} and~\Thref{thm:polarised-sampled}, provide a theoretical explanation of the the discrepancies between the averaged Mutual Information (MI)
        and total correlation (TC) scores of mean and sampled representations for passive variables components.
        \item Verify that the correlation of mean representations increases with
        the number of passive and mixed variables.~This will allow us to explore how the number of passive variables evolves with stronger regularisation and whether this number can explain the discrepancies in total correlation and averaged mutual information that were reported by \cite{Locatello2019a}.
        \item Empirically verify the results of step (ii) by comparing the impact of passive and mixed variables on averaged mutual information and total correlation scores for mean and sampled representations.~By separately comparing the scores of every combination of variable
        type, we can further attribute the discrepancies to a specific type of variable, or a combination of variables.
    \end{enumerate}
    These steps are implemented in \Foursecrefs{subsec:step1}{subsec:step2}{subsec:step3}{subsec:step4}, respectively.

    \paragraph{Empirical setup}
    We based our implementation on \texttt{disentanglement\_lib}~\citep{Locatello2019a}, using the same datasets as the authors:
    dSprites~\citep{Higgins2017}, smallNorb~\citep{LeCun2004}, cars3D~\citep{Reed2015}, and the alternative versions of
    dSprites~\citep{Locatello2019a} color-dSprites, Scream-dSprites and Noisy-dSprites. We relied on the 9000
    pre-trained models released by~\citet{Locatello2019a}, corresponding to the 5 VAE architectures described in
    \Secref{sec:background}, over the 6 datasets mentioned above, with 6 different regularisation strengths for each method
    and 50 seeds per (dataset, method, regularisation) triplet.

    \subsection{Identifying variable types}\label{subsec:step1}
    Using~\Propref{prop:polarised-var} and defining a threshold $\alpha \geq 0$, one can easily classify the type of
    variables stored at each index $j \in d$ based on the variance representations:
    \begin{itemize}
        \item \textbf{Passive variable}: $\bar{\rvsigma}_j = 1 \pm \alpha \text{ and } Var(\rvsigma_j) = 0 \pm \alpha$,
        \item \textbf{Active variable}: $\bar{\rvsigma}_j = 0 \pm \alpha \text{ and } Var(\rvsigma_j) = 0 \pm \alpha$,
        \item \textbf{Mixed variable}: any variable not classified as active or passive.
    \end{itemize}
    In all our experiments, $\alpha$ is set to $0.1$.
    As stated above, any variable that does not satisfy both conditions required for active or passive variables becomes a mixed variable.\par
    Note that the indexes obtained with this method are used in the same way to identify the variable types of mean and sampled representations.
    For example, if our procedure determines that index $j=1$ of the variance representation corresponds to a passive variable, the variable at index $j=1$ of the mean and sampled representations will be considered as passive.

    \paragraph{Sanity check} To verify that our thresholds are valid and that the variable types have been mapped
    correctly, we compare the scores of a classifier trained on the whole representations with those of classifiers trained on
    every combination of variable types.~Similarly to~\citet{Locatello2019a}, the classification models are
    logistic regressions (LRs) that predict the labels of the dataset from which the representation was learned.\par
    We trained the LRs for 300000 steps on 10000 data examples for each variable combination and computed the average accuracy
    over 5000 test examples.~The LRs are cross-validated with 10 different regularisation strengths and 5 folds.\par

    If our procedure to identify active, mixed and passive variables is correct, we should expect the following. Active variables should contribute the most
    to the predictions and give results close to those obtained by the full representation.~Mixed variables should contribute
    less and have a much lower score but together with the active representation, they should provide the same score as the
    full representation.~Passive variables should not contribute at all and ought to provide results close to a random classifier.\par
    \begin{figure}[h!]
        \centering
        \begin{subfigure}{.5\textwidth}
            \centering
            \includegraphics[width=\linewidth,keepaspectratio]{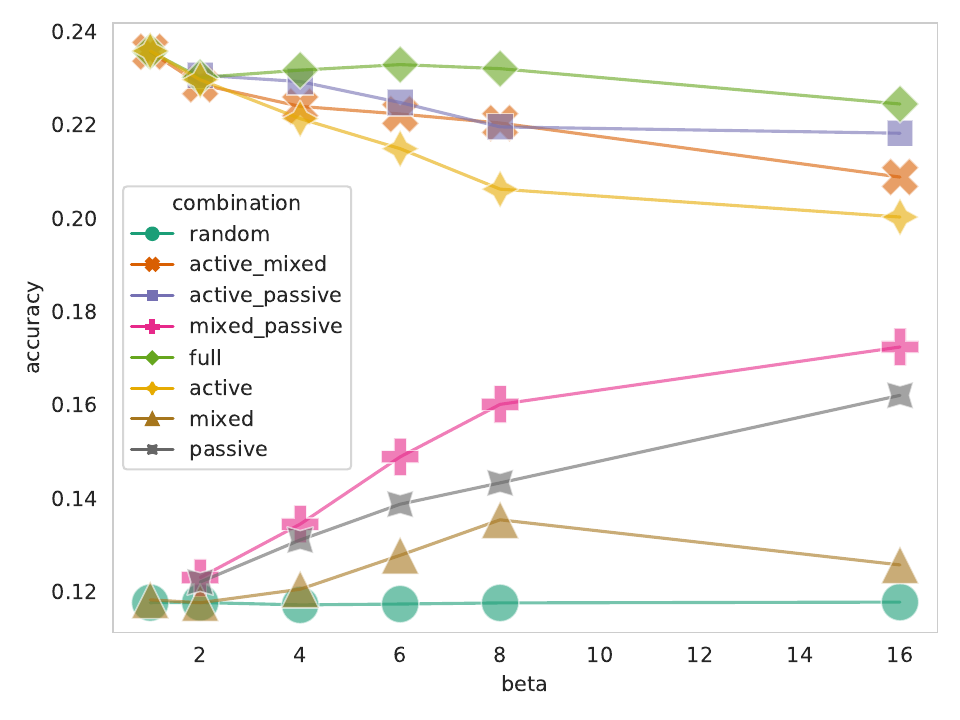}
            \caption{Mean representation}\label{subfig:threshold-check-mean}
        \end{subfigure}\begin{subfigure}{.5\textwidth}
                           \centering
                           \includegraphics[width=\linewidth,keepaspectratio]{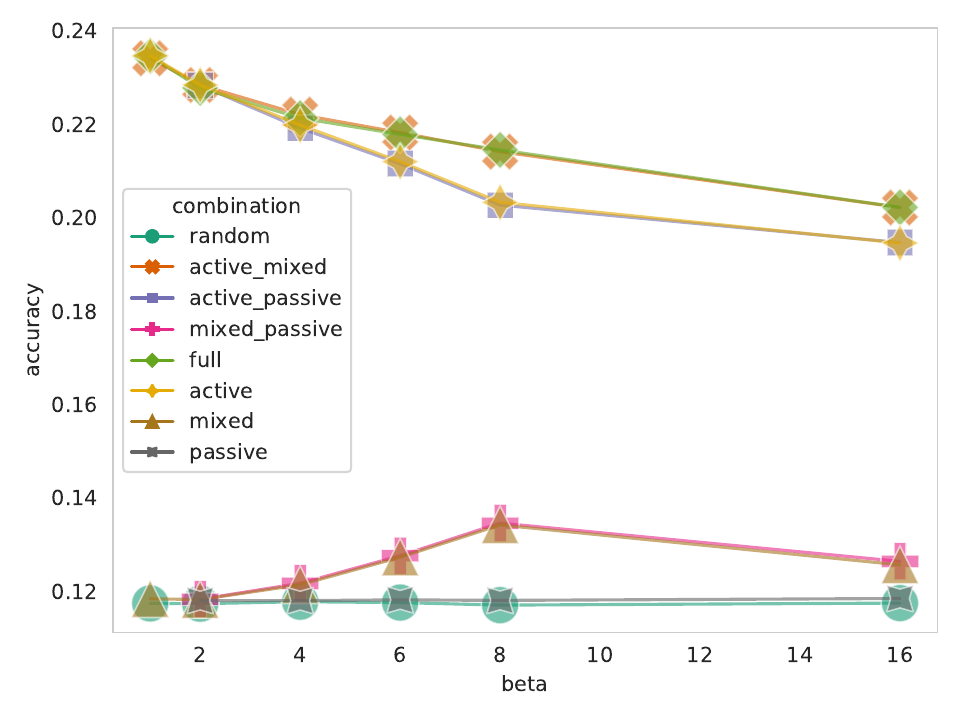}
                           \caption{Sampled representation}\label{subfig:threshold-check-sampled}
        \end{subfigure}
        \caption{Average test accuracy of a logistic regression trained on the mean and sampled representations learned by a $\beta$-VAE trained on dSprites.
        Each figure shows the results obtained using the full representations and combinations of different variable types. This is also compared to a random
        classifier picking uniformly from the possible labels.}
        \label{fig:threshold-check}
    \end{figure}

    In \Figref{subfig:threshold-check-sampled}, we can see that the results obtained for $\beta$-VAE trained on dSprites using sampled
    representations are exactly as expected.~The passive variables gave equivalent results to a random
    classifier, while mixed variables performed slightly better, which indicates that we correctly identify them.
    The score obtained with active and mixed representations is the same as the full score in sampled representations, while
    active variables alone or with passive variables performed a bit worse, which confirms that mixed and active
    variables have also been identified correctly.\par

    Interestingly, the results obtained for mean representations in \Figref{subfig:threshold-check-mean} show that passive
    variables are becoming more informative as the regularisation strength increases.
    While the score obtained with active variables is still closer to the full representation score, we can see that in opposition to sampled representations, passive and active
    variables perform better than mixed and active variables.~Thus, \textit{despite being close to zero and having a
    very low variance, passive variables of the mean representation seem to capture some information about the data}.
    ~Note that this result does not indicate a problem in the detection of the passive variables: if we had incorrectly identified any active or
    mixed variables as passive, they would still convey some information in the sampled representation.
    As seen in \Figref{subfig:threshold-check-sampled}, it is not the case here as their performance is equivalent to the performance of a random classifier.

\begin{figure}[h!]
        \centering
        \begin{subfigure}{.5\textwidth}
            \centering
            \includegraphics[width=\linewidth,keepaspectratio]{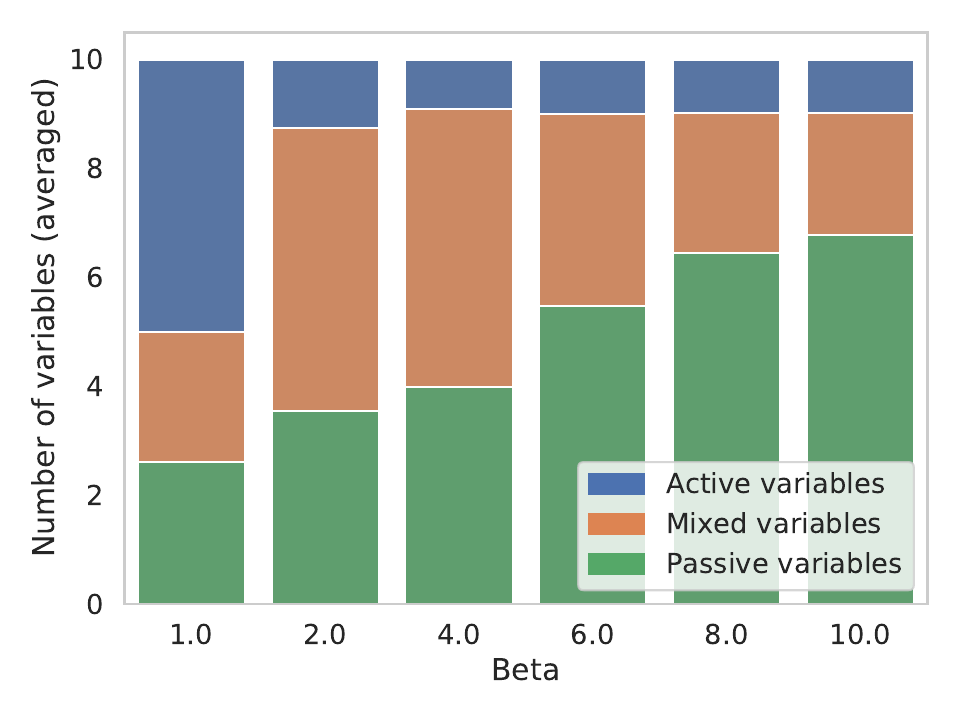}
            \caption{$\beta$-TC VAE trained on smallNorb.}
            \label{fig:passive-var-evol-1}
        \end{subfigure}\begin{subfigure}{.5\textwidth}
                           \centering
                           \includegraphics[width=\linewidth,keepaspectratio]{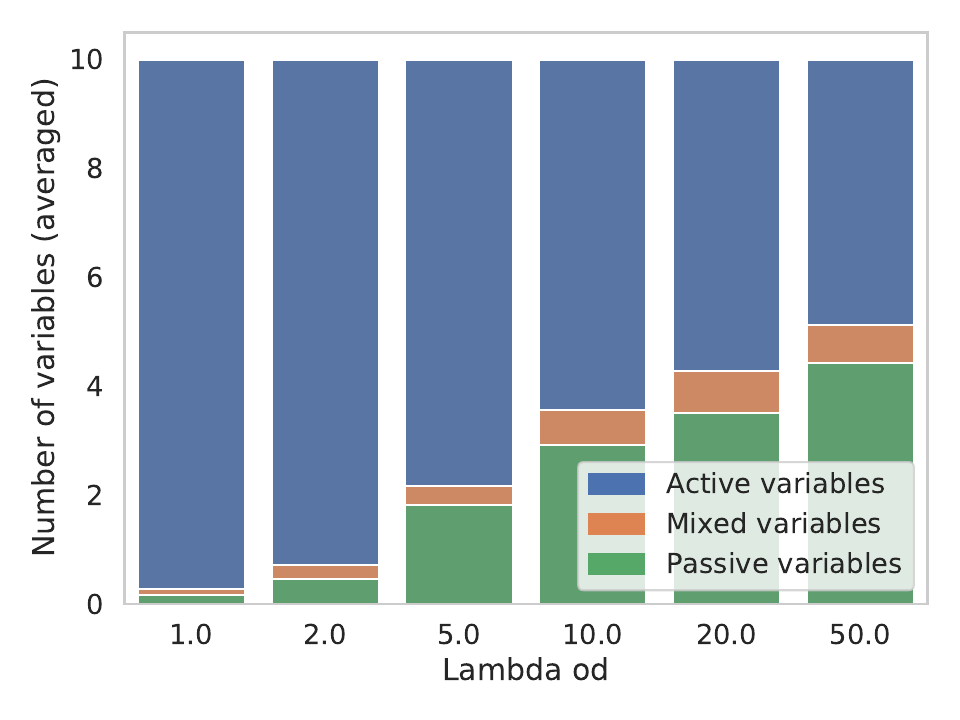}
                           \caption{DIP-VAE II trained on dSprites.}
                           \label{fig:passive-var-evol2}
        \end{subfigure}
        \caption{Number of passive, mixed and active variables with increased regularisation strength averaged over 50 runs for each regularisation value.}
        \label{fig:passive-var-evol}
    \end{figure}

    As discussed in \Secref{sec:polarised-regime}, we can also see in \Figref{fig:passive-var-evol} that, as expected, the number of passive and mixed variables increases with the regularisation strength.
    The number of passive variables at the lowest regularisation strength is generally close to zero for all datasets and all models,
    except annealed-VAE, whose special case will be discussed in \Appref{sec:annealed-vae}.

    To summarise, given that the behaviour of the passive and active variables observed in \Twofigref{fig:threshold-check}{fig:passive-var-evol}
    is consistent with the findings of~\citet{Rolinek2019} and~\citet{Dai2020} regarding the polarised regime, we assume that our method to determine the type of variables
    is valid, and our thresholds properly set.~We can thus proceed to the next steps of our experiment.

    \subsection{Metrics used to assess mean and sampled representations}\label{subsec:step2}
    To be consistent with the existing literature~\citep{Locatello2019a}, we compared the mean and sampled representations using total correlation and averaged mutual information
    scores.~We also used effective rank~\citep{Roy2007} as a complementary measure. Those metrics are measuring slightly different things and have different
    limitations that we will detail below.

    \subsubsection{Total correlation}
    The total correlation~\citep{Watanabe1960} is a measure of the amount of
    information shared between multiple latent variables.~It is measured as the KL divergence between the joint
    distribution and the product of its marginal distributions.
    More formally, given a latent representation $\rvr = \rvr(\rvx; \phi)$:
    \begin{equation}
        TC(\rvr) = \klbigg{p(\rvr)}{\prod_{j=1}^d p(\rvr_j)}\label{eq:tc}.
    \end{equation}
    In their experiment,~\citet{Locatello2019a} assumed that $\rvr$ (either the mean or sampled representation) was multivariate Gaussian
    with a mean of zero. Thus, given the mean $\bar{\rvr}$ and covariance $\Cov[\rvr]$ of $\rvr$,~\Eqref{eq:tc} takes the following form:
    \begin{equation}
        TC(\rvr) = \klbigg{\N(\bar{\rvr}, \Cov[\rvr])}{\prod_{j=1}^d \N(\bar{\rvr}_j, \Cov[\rvr]_{jj})}\label{eq:tc-gaussian-raw},
    \end{equation}
    which simplifies to
    \begin{equation}
        TC(\rvr) = \frac{1}{2}\bigg(\sum_{j=1}^d(\log \Cov[\rvr]_{jj}) - \log{\det(\Cov[\rvr])} \bigg)\label{eq:tc-gaussian}.
    \end{equation}
    The transition from \Eqref{eq:tc-gaussian-raw} to \Eqref{eq:tc-gaussian} can be found in \Appref{subsec:calc-kl-tc}.
    From~\Eqref{eq:tc-gaussian}, we then obtain~\Thref{thm:tc-pv}, as proved in~\Appref{subsec:proof-tc-pv}.
    \begin{theorem}[Impact of passive variables on $TC(\rvz)$]
        \label{thm:tc-pv}
        The total correlation of sampled representations is not modified by passive variables.
    \end{theorem}

    \paragraph{Limitations}
    According to~\Thref{thm:polarised-sampled}, while the passive variables of the sampled representations follow standard Gaussian distributions,
    other variables of mean and sampled representations, especially mixed variables, may not be Gaussian.~Moreover, the distributions of the latent variables may not be jointly Gaussian.
    As such, TC may provide inaccurate results.~For this reason, it is important to use complementary metrics.

    \subsubsection{Averaged mutual information}
    Mutual information is a measure of the information shared by two latent variables~\citep{Cover1999}.
    Similarly to TC, which is a generalisation of mutual information to the multivariate case, it is measured as the KL
    divergence between the joint distribution and the product of its marginal distributions.
    ~That is, given two latent factors $\rvr_1$ and $\rvr_2$:
    \begin{equation}
        MI(\rvr_1, \rvr_2) = \klbig{q(\rvr_1, \rvr_2)}{q(\rvr_1)q(\rvr_2)}\label{eq:mi}.
    \end{equation}
    In their experiment, \citet{Locatello2019a} calculated the mutual information over discretised values using:
    \begin{equation}
        MI(\rvr_1, \rvr_2) = \sum_{i=1}^{|\mU|}\sum_{j=1}^{|\mV|} \frac{|\mU_i \cap \mV_j|}{n} \log \frac{n |\mU_i \cap \mV_j|}{|\mU_i||\mV_j|}\label{eq:discrete-mi},
    \end{equation}
    where $\mU$ and $\mV$ are the bins of $\rvr_1$ and $\rvr_2$ respectively, $n$ is the number of samples,
    and $|\cdot|$ denotes the cardinality.
    They then used the averaged mutual information over all the latent factors:
    \begin{align}
        MI_{avg}(\rvr) &= \frac{1}{k^2 - k}\sum_{i=1}^k \sum_{j\neq i} MI(\rvr_i, \rvr_j) \label{eq:avg-discrete-mi},
    \end{align}
    where $k$ is the dimensionality of the latent representation $\rvr$.
    \Thref{thm:mi-pv} follows, as proved in~\Appref{subsec:proof-mi-pv}.

    \begin{theorem}[Impact of passive variables on $MI_{avg}(\rvz)$]
        \label{thm:mi-pv}
        The averaged mutual information of sampled representations decreases with the number of passive variables.
    \end{theorem}

    \paragraph{Limitations}
    As it is using discretised values, averaged MI does not have the downside of TC regarding the Gaussian assumption.
    However, because MI is averaged, it may diminish the strong relationships between two variables if the other MI scores
    are close to zero, as seen in~\Thref{thm:mi-pv}.

    \subsubsection{Effective rank}
    The effective rank of a matrix~\citep{Roy2007} is a real-valued generalisation of the integer-valued rank of a matrix.
    More formally, let us consider a matrix $\mA \in \R^{m \times n}$. Its singular value decomposition is $\mA = \mU\mS\mV^T$ where
    $\mU \in \R^{m \times m}$ and $\mV \in \R^{n \times n}$ are orthogonal matrices, and $\mS \in \R^{m \times n}$ is a
    rectangular diagonal matrix containing the $l$ singular values
    $s_1 \geqslant s_2 \geqslant \cdots \geqslant s_l \geqslant 0$ where $l = \min(m,n)$.
    The singular value distribution is given by:
    \begin{equation*}
        p_k = \frac{s_k}{\sum_{i=1}^{l} s_i} \text{ for } k = 1, 2, \cdots, l.
    \end{equation*}
    The effective rank is then defined as:
    \begin{equation*}
        \text{erank}(\mA) =  \exp\big( \text{H}(p_1, p_2, \cdots, p_l) \big),
    \end{equation*}
    where $\text{H}(p_1, p_2, \cdots, p_l)$ is the spectral entropy~\citep{Campbell1960,Yang2005}:
    \begin{equation*}
        \text{H}(p_1, p_2, \cdots, p_l) = - \sum_{k=1}^{l} p_k \log p_k.
    \end{equation*}

    The effective rank is generally more informative than the rank as it can take all possible values in the interval $[1, l]$, whereas the rank is limited to integer values in the set $\{1, 2, . . . , l\}$.~Consider, for example, a Gaussian distribution of
    dimension two whose variables are highly correlated.~The first singular value will dominate while the second will be close to zero.
    As both singular values are higher than zero, the matrix rank will be two.
    However, because the second singular value is very low, its effective rank will be only slightly above one. Overall, the effective rank tells us more about the data than the integer-valued rank.\par

    \paragraph{Limitations}
    While the effective rank does not have the weaknesses of TC regarding the Gaussian assumption nor averaged MI regarding the
    dilution of a strong relationship when dimensionality is high, it would not make sense to measure it separately for
    each type of variables as the significance of the dimensions would be relative to the subset considered.~Thus, effective rank is
    only included in \Secref{subsec:step3}, where we use it to analyse all variables jointly.

    \subsection{Preliminary observations}\label{subsec:step3}
    As we have established in \Secref{sec:polarised-regime}, the active variables of mean and sampled representations should
    be similarly correlated.~Hence, while \citet{Locatello2019a} concluded that uncorrelated sampled representations did
    not guarantee uncorrelated mean representations, we argue that active variables, which encode the most information,
    should have similar correlation in mean and sampled representations.
    ~We thus suggest that the increased correlation of mean representations may be due to mixed and passive variables.
    If the passive variables are, as hypothesised, responsible for the higher correlation of mean representations,
    one should expect the effective rank to be close to the total number of latent variables minus the passive ones.

\begin{figure}[h!]
        \centering
        \begin{subfigure}{.33\textwidth}
            \centering
            \includegraphics[width=\linewidth,keepaspectratio]{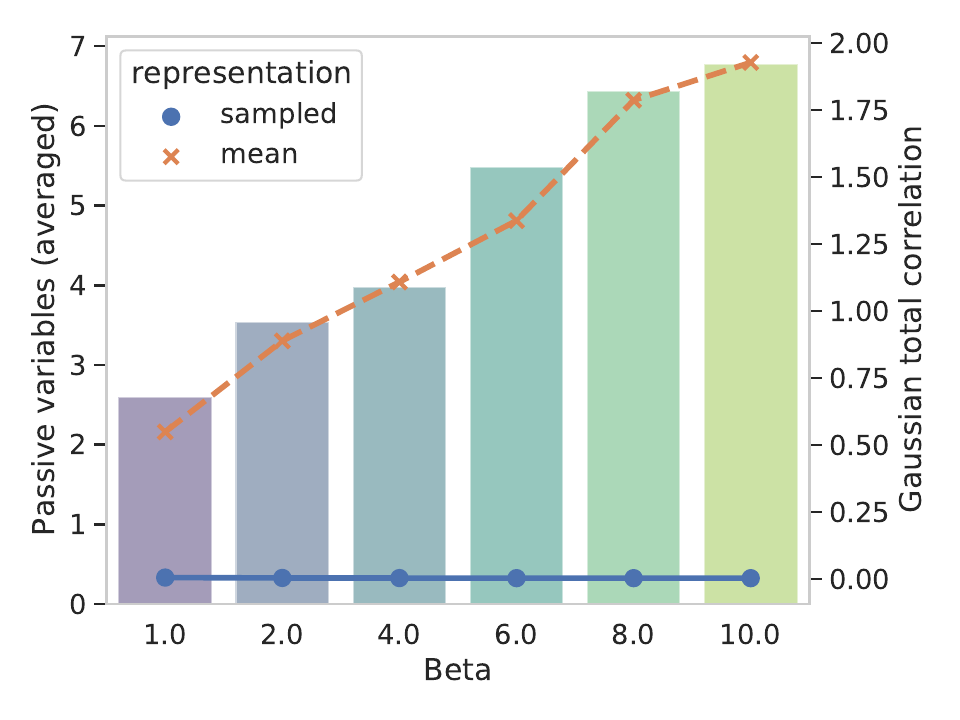}
            \caption{Total correlation}
            \label{fig:TC-PV1}
        \end{subfigure}\begin{subfigure}{.33\textwidth}
                           \centering
                           \includegraphics[width=\linewidth]{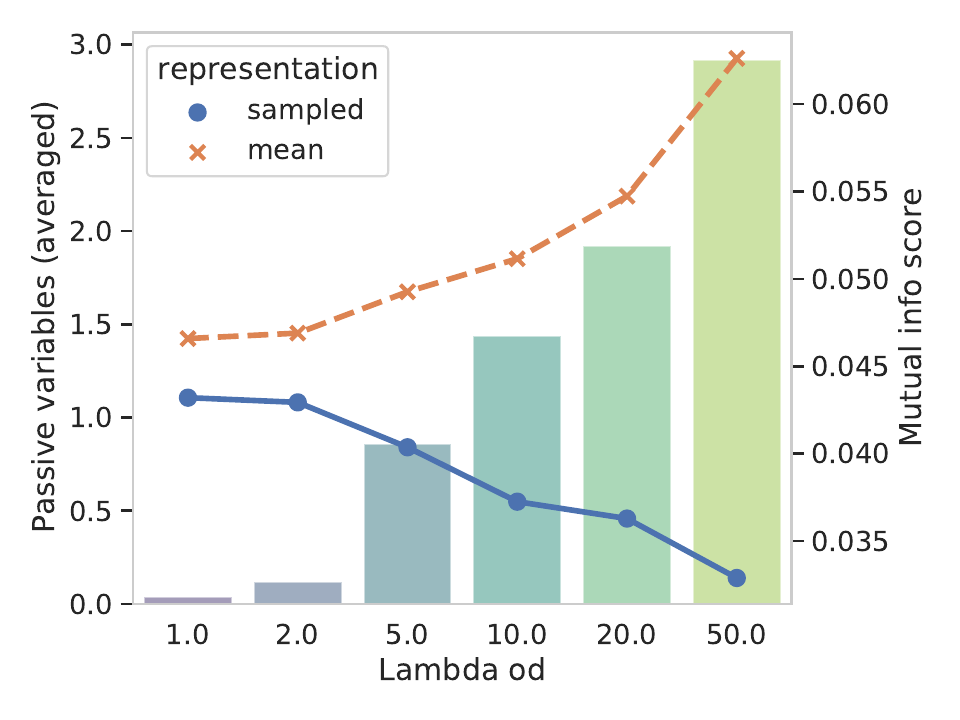}
                           \caption{Mutual information}
                           \label{fig:MI-PV1}
        \end{subfigure}\begin{subfigure}{.33\textwidth}
                           \centering
                           \includegraphics[width=\linewidth]{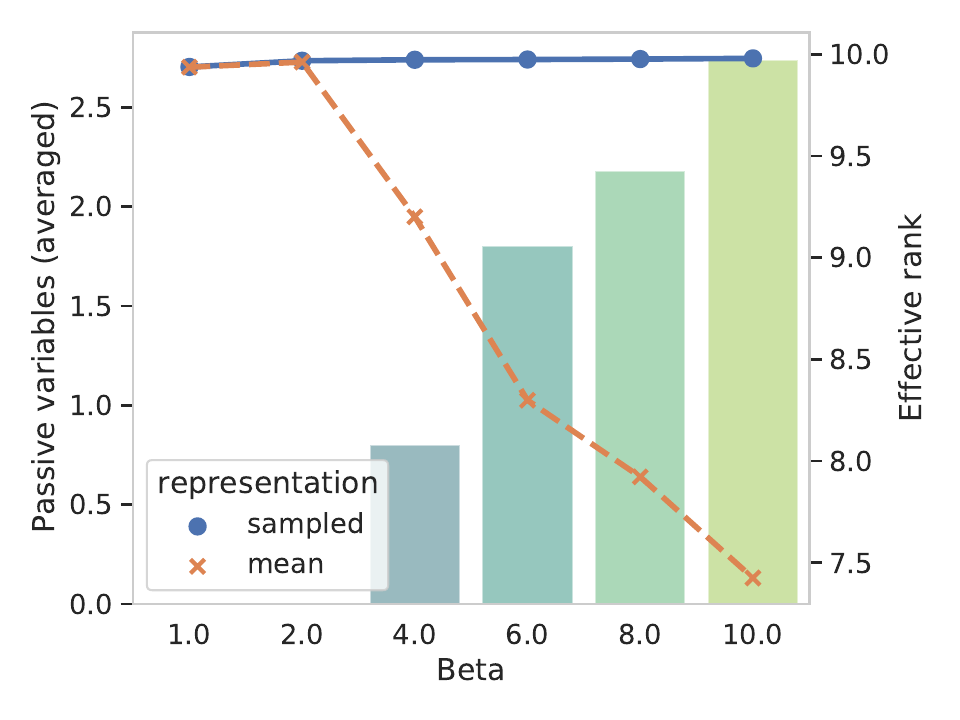}
                           \caption{Effective rank}
                           \label{fig:ER-PV1}
        \end{subfigure}\caption{Relationship between the number of passive variables and the total correlation, mutual information, and effective rank scores of mean and sampled representations.
        Figures (a), (b) and (c) use the representations learned by $\beta$-TC VAE trained on smallNorb, DIP-VAE II trained on Cars3D, and $\beta$-TC VAE trained on colour dSprites, respectively.
        Lines indicate the metric scores of the representation and the bars the average number of passive variables.}
        \label{fig:metric-pv}
    \end{figure}

    Our preliminary observations are consistent with this hypothesis.
    Specifically, in sampled representations,~\Twofigref{fig:TC-PV1}{fig:MI-PV1} show that TC does not change when we increased the number of passive variables and MI decreases,
    as described in~\Twothmrefs{thm:tc-pv}{thm:mi-pv}.
    Moreover,~\Figref{fig:TC-PV1} shows that the discrepancies between the total correlation
    scores of the mean and sampled representations are increasing with the number of passive variables, and we can observe the same
    trend for mutual information in~\Figref{fig:MI-PV1}.
    The effective rank of mean representations is, as shown in \Figref{fig:ER-PV1}, close to the total number of latent
    variables minus the passive ones.~One can also notice that in sampled representations, the effective rank is close to
    the total number of variables as the passive variables are replaced by uncorrelated samples from $\mathcal{N}(0, 1)$, and those uncorrelated variables cannot reduce the effective rank any more.
    Interestingly, in addition to showing an increased correlation of passive variables, the effective rank is also providing
    further confirmation that passive variables are correctly identified in~\Propref{prop:polarised-var}
    and \Secref{subsec:step1}.

    \subsection{Impact of passive and mixed variables on averaged mutual information and total correlation}\label{subsec:step4}
    To further validate our hypothesis that only mixed and passive variables are responsible for the increased correlation
    of mean representations, we will compare TC and averaged MI scores with and without mixed and passive variables.
    Thus, we will be able to determine whether we can have an increased correlation between any dimensions of the mean
    representations, as initially inferred by \citet{Locatello2019a}, or only between a specific subset corresponding to mixed and
    passive variables.

    As we have mapped each index of the mean and sampled representation to a specific variable type in~\Secref{subsec:step1}, we can now
    assess the impact of mixed and passive variables on TC and averaged MI by comparing the scores with and without them.
    We first calculate TC and averaged MI using the full representation and compare these scores with those obtained without passive variables
    and with only active variables.

    \begin{figure}[ht!]
        \centering
        \begin{subfigure}{.33\textwidth}
            \centering
            \includegraphics[width=\linewidth]{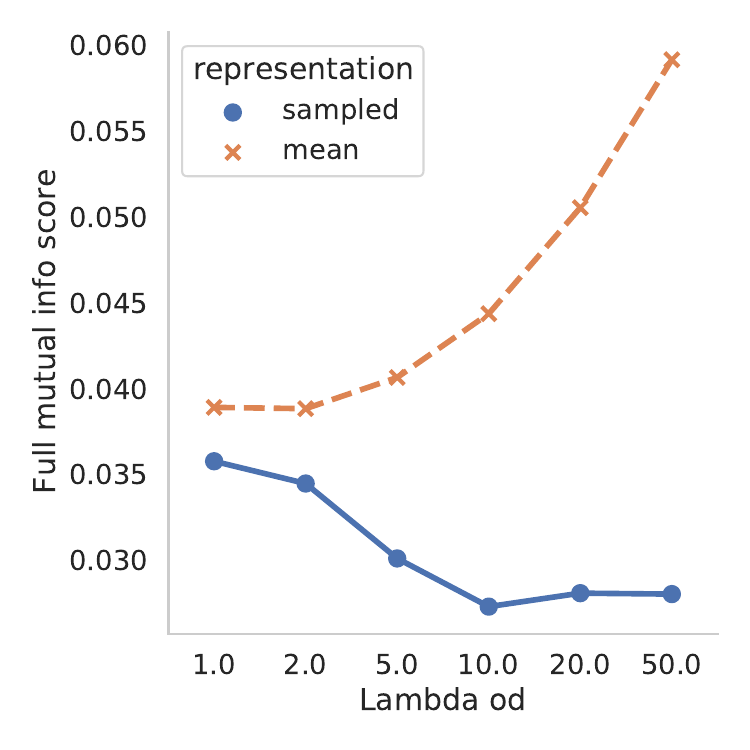}
            \caption{Full representation}
            \label{fig:MI-full}
        \end{subfigure}\begin{subfigure}{.33\textwidth}
                           \centering
                           \includegraphics[width=\linewidth]{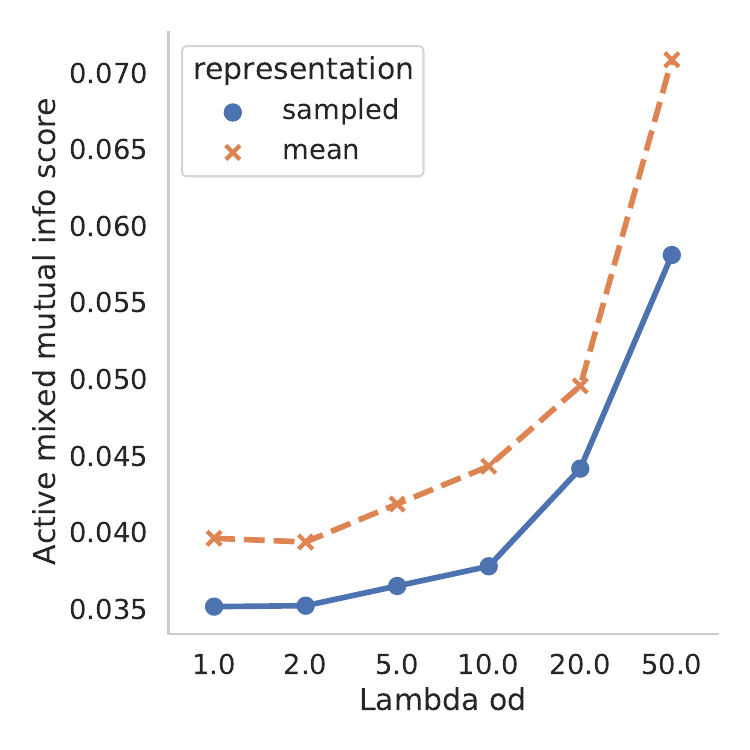}
                           \caption{Mixed and active variables}
                           \label{fig:MI-active-mixed}
        \end{subfigure}\begin{subfigure}{.33\textwidth}
                           \centering
                           \includegraphics[width=\linewidth]{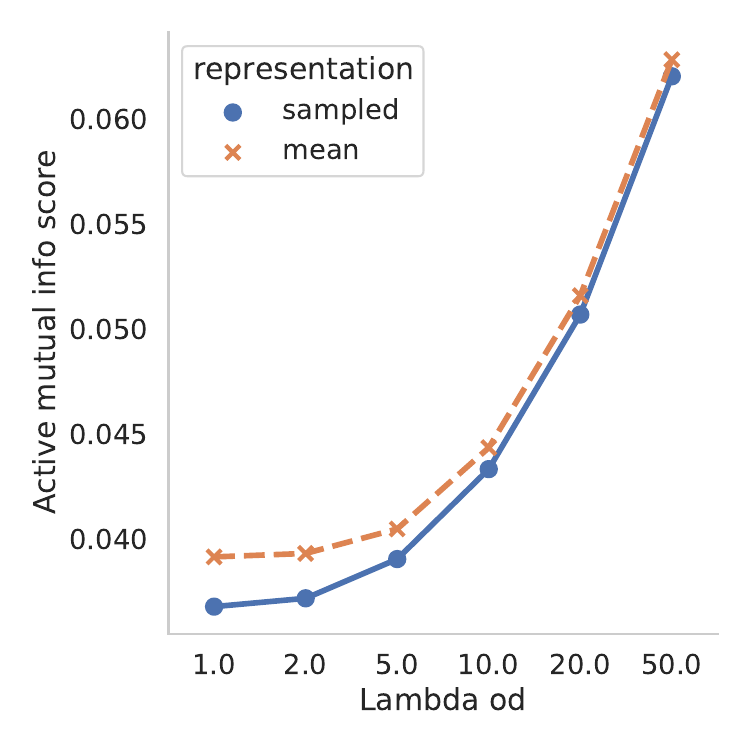}
                           \caption{Active variables}
                           \label{fig:MI-active}
        \end{subfigure}
        \caption{Comparison of the averaged mutual information scores of the mean and sampled representations of DIP-VAE II trained on dSprites.
        In Figure (a), the number of passive variables increases with $\lambda$ leading to lower averaged MI scores for sampled representations.
        The mean and sampled representation scores become more similar once passive variables are removed in Figures (b) and (c).}
        \label{fig:MI-comb}
    \end{figure}
    \begin{figure}[ht!]
        \centering
        \begin{subfigure}{.33\textwidth}
            \centering
            \includegraphics[width=\linewidth]{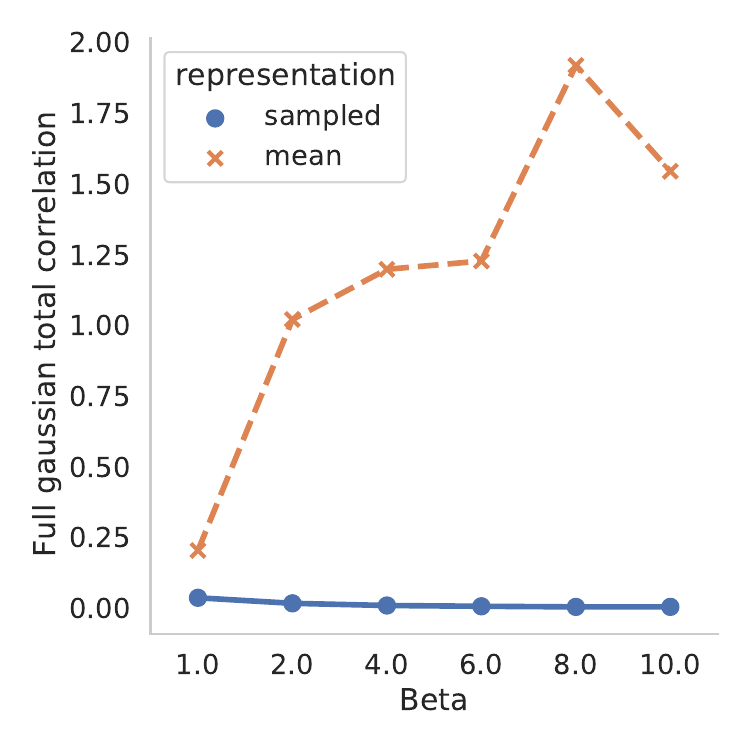}
            \caption{Full representation}
            \label{fig:TC-full}
        \end{subfigure}\begin{subfigure}{.33\textwidth}
                           \centering
                           \includegraphics[width=\linewidth]{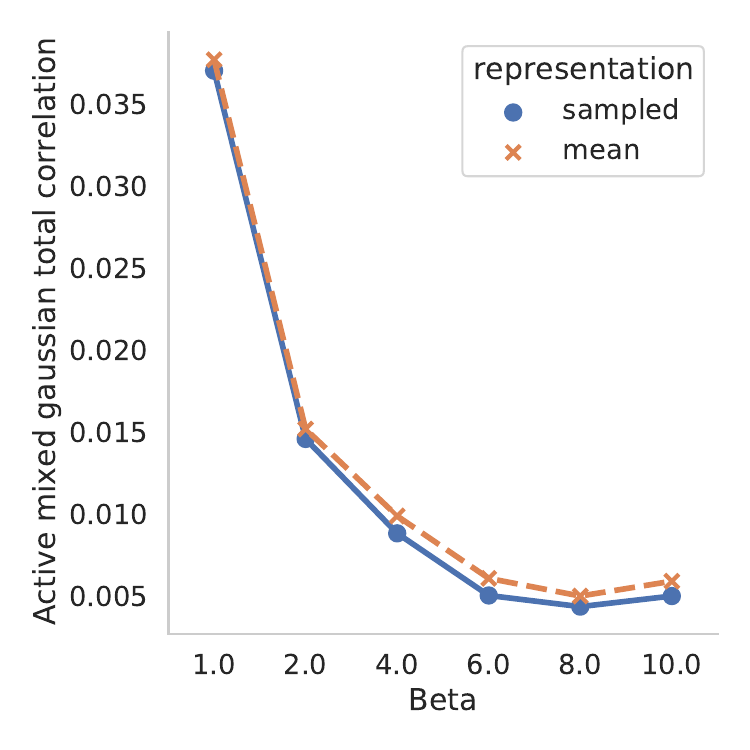}
                           \caption{Mixed and active variables}
                           \label{fig:TC-active-mixed}
        \end{subfigure}\begin{subfigure}{.33\textwidth}
                           \centering
                           \includegraphics[width=\linewidth]{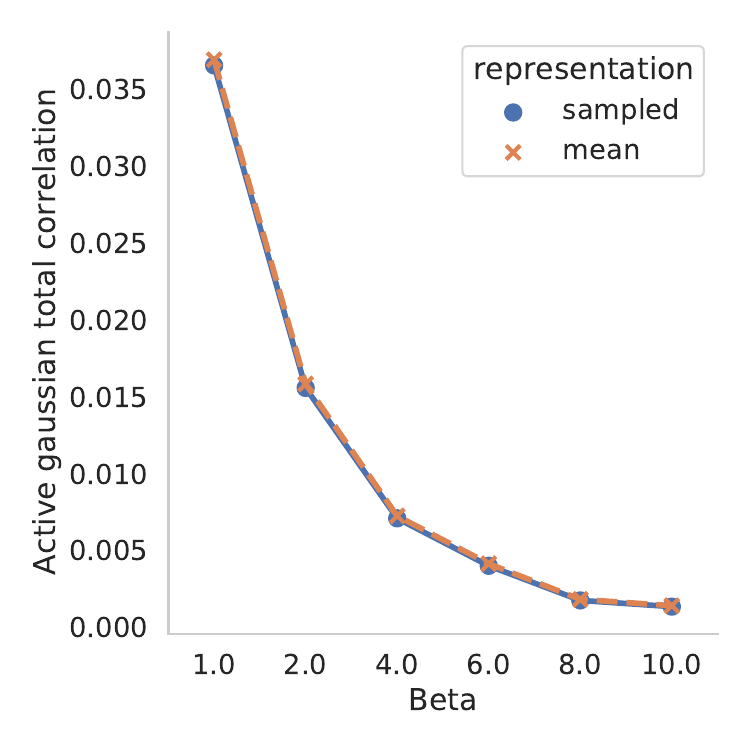}
                           \caption{Active variables}
                           \label{fig:TC-active}
        \end{subfigure}
        \caption{Comparison of the total correlation scores of the mean and sampled representations of $\beta$-TC VAE trained on noisy dSprites.
        In Figure (a), the number of passive variables increases with $\beta$ leading to lower TC scores for sampled representations.
        The mean and sampled representation scores become more similar once passive variables are removed in Figures (b) and (c).}
        \label{fig:TC-comb}
    \end{figure}

    \paragraph{Mutual information and total correlation of mean representations}
    In~\Twofigref{fig:MI-full}{fig:TC-full}, we can see that the mixed and passive variables are
    raising the TC and averaged MI scores of the mean representations (the dashed orange curves).~Indeed, when these variables are removed, the TC and averaged MI scores of the mean and sampled
    representations are quite similar, as observed in~\Twofigref{fig:MI-active-mixed}{fig:MI-active}, and~\Twofigref{fig:TC-active-mixed}{fig:TC-active}.
    While the mixed variables impact the score to a small extent, the passive variables
    lead to a dramatic score increase, especially for TC.~These observations, thus, show that \textit{active variables are as disentangled in mean as in sampled representations,
        and passive variables of the mean representations should have strong correlations with other variables.}
    We, therefore, examine the correlation of passive variables in~\Secref{sec:passive-variables} to better understand with which type
    of variable they are correlated and how this correlation emerges.~Even before we present this analysis, we should recall that passive variables
    of the mean representations performed better on downstream tasks than their sampled counterpart, as observed in~\Secref{subsec:step1}.
    Thus, to convey useful information, the passive variables have to be correlated with known informative variables.~As a result, one should
    expect the passive variables in the mean representations to be correlated with active ones.~This is the subject of our investigation in
    \Secref{sec:passive-variables}.
    \section[Origin of the correlation of passive variables]{Where does the correlation of passive variables in the mean representation come from?}\label{sec:passive-variables}
\begin{figure}[h!]
        \centering
        \begin{minipage}{.45\textwidth}
            \centering
            \includegraphics[width=\linewidth]{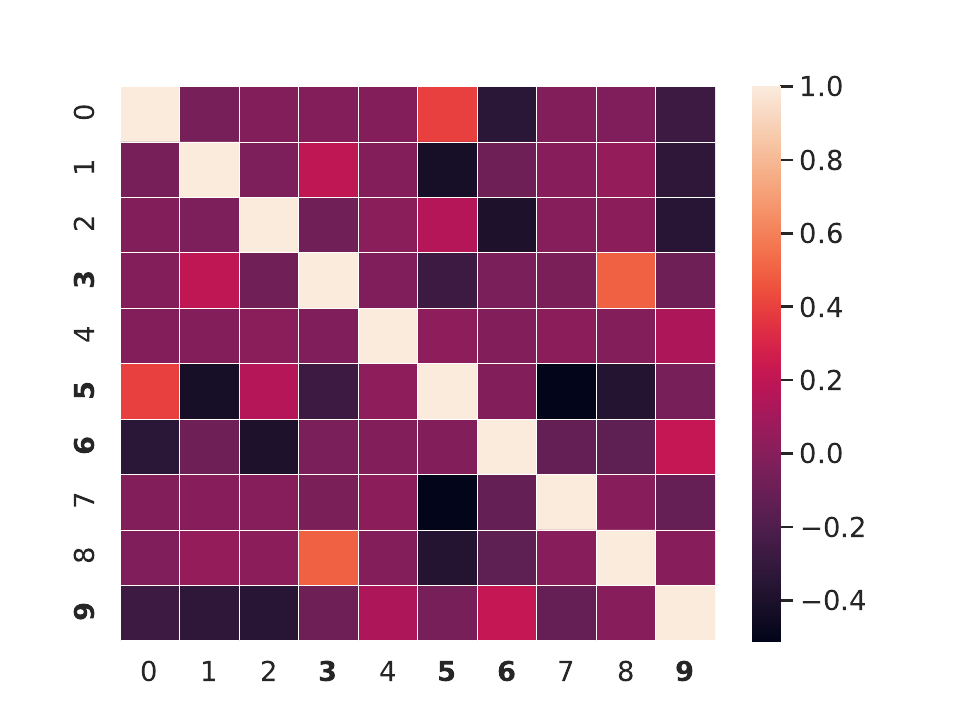}
            \caption{Correlation of the passive variables of $\beta$-VAE trained on colour dSprites with $\beta=16$ with other variables.
            The passive variables are at indexes 3, 5, 6, and 9. Several correlations of the passive variables clearly stand out.}
            \label{fig:pv-correlation}
        \end{minipage}\hfill
        \begin{minipage}{.45\textwidth}
            \centering
            \includegraphics[width=\linewidth]{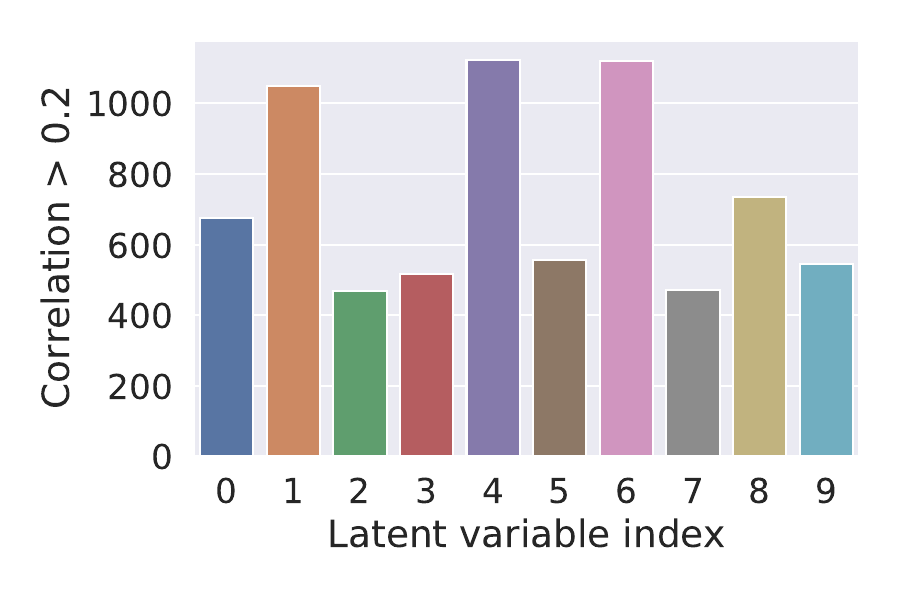}
            \caption{Number of times where the absolute value of the correlation of each latent variable with another
            variable in the mean representation was above 0.2 in the 300 snapshots recorded during the training of a
                $\beta$-VAE with $\beta=8$ on dSprites.
                The passive variables are at indexes 1, 4, and 6, and are the most often correlated variables.}
            \label{fig:count-correlation}
        \end{minipage}
    \end{figure}

    As the passive variables of mean representations lead to higher TC and averaged MI scores, they should exhibit some
    correlation with other variables.~This can be seen empirically in~\Figref{fig:pv-correlation} where the correlation between
    each latent variable of a $\beta$-VAE trained with $\beta = 16$ on colour dSprites shows that, indeed, the passive variables tend to have
    strong correlation scores with one or more variables, which is generally not the case for other variable types.
    This result should not be surprising as VAEs optimise passive variables to be close to zero with low variance,
    but not to be uncorrelated.~However, one can ponder on the origin of these correlations.~Are they present from the
    beginning of the learning process and lead these variables to become passive, or are the correlations the consequence of the variables being passive?\par

\begin{figure}[h!]
        \centering
        \includegraphics[width=\linewidth]{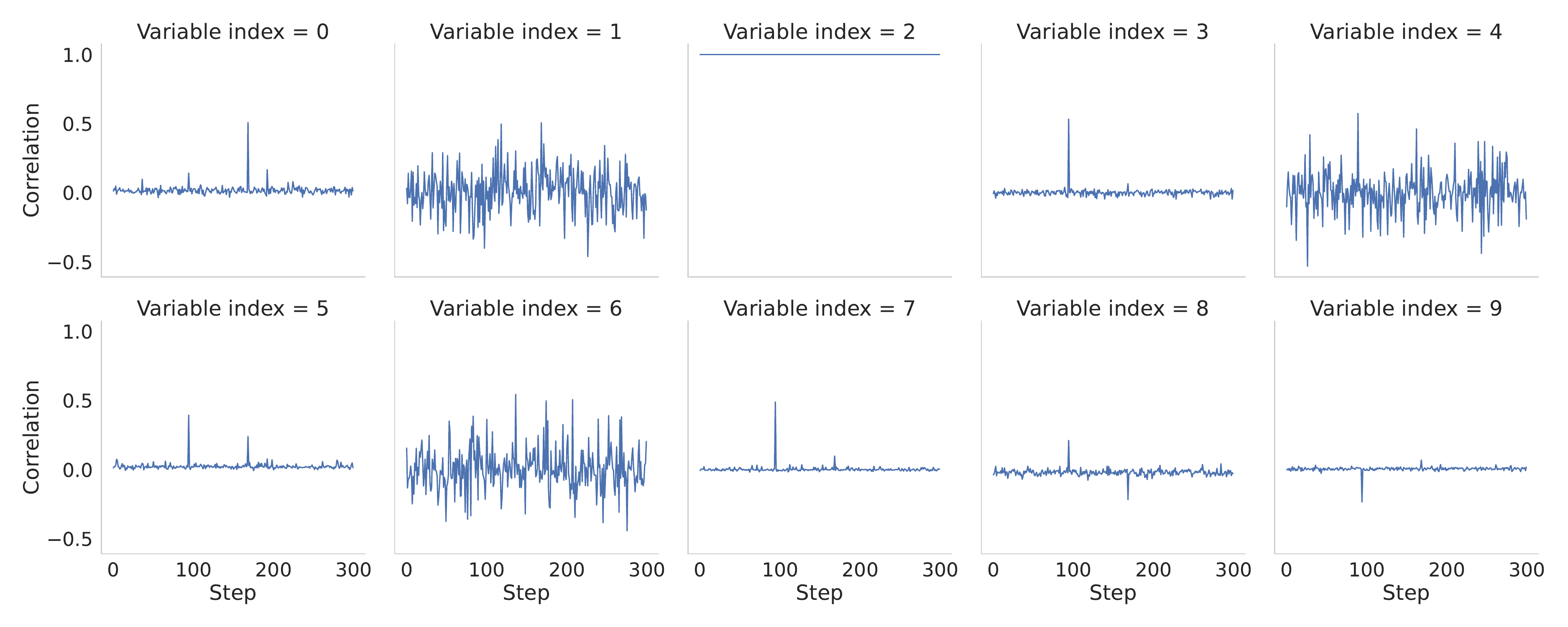}
        \caption{The correlation scores of the active variable at index 2 of the mean representation with all the other
        variables during the 300K training steps of a $\beta$-VAE with $\beta=8$ trained on dSprites.
        We can see an increased correlation with all the passive variables (indexes 1, 4, and 6).}
        \label{fig:corr-steps}
    \end{figure}

    To gain some insights into this question, we trained
    a $\beta$-VAE with $\beta=8$ on dSprites for 300K steps and saved a snapshot of the model parameters every 1000 steps and observed the
    evolution of the latent representations.~Using the same technique as in~\Secref{sec:truncation-experiment},
    after the model has been trained (i.e.~after 300K training steps), we determined
    that the variables 1, 4, and 6 were passive, the variables 0, 3 and 8 mixed, and the remaining ones active.
    In~\Figref{fig:count-correlation}, we can see that the correlation score of the passive variables is more often above
    0.2 than the scores of the active variables across the 300 snapshots recorded during the training process.
    This highlights the important correlation scores of the passive variables in most of the training steps.
    \Figref{fig:corr-steps} shows that the correlation between passive variables and other variables varies significantly during
    the training process and can be relatively high, while the correlation scores of active variables remain very low.~These high
    correlations are consistent with the observations of~\Secref{subsec:step1} where the logistic regression had better accuracy with
    passive variables than with mixed ones.
    While a more in-depth study of the learning dynamics of VAEs would be needed to provide a complete explanation of this
    phenomenon, the frequent changes of the correlation scores makes it likely to be an inherent property of the neural network training process.

    \section{Conclusion}
    In their study,~\citet{Locatello2019a} have reported that mean representations seemed to be more correlated than sampled
    ones in a large number of experiments.~They concluded that enforcing uncorrelated sampled representations may not be sufficient to obtain uncorrelated
    mean representations.
    By extending the definition of the polarised regime to mean representations,
    we have shown that while aiming to optimise uncorrelated sampled representations is not sufficient to guarantee
    completely uncorrelated mean representations, it is sufficient to obtain uncorrelated \textit{active variables} in mean
    representations.~We thus hypothesised that the increasing discrepancies between the two representations should only be
    attributed to a subset of variables: the passive ones.~This hypothesis was consistent across different levels of
    regularisation and has been further confirmed by empirical observations showing an increased correlation of passive
    variables.
    By considering the latent representations over the whole dataset, we have also introduced mixed variables,
    a type of variable that can either be active or passive depending on the input example provided. Our empirical results confirmed the existence and importance of such variables.

    \paragraph{Should we use mean representations for downstream tasks?} One of the concerns that was raised by the findings of
    \citet{Locatello2019a} was that mean representations, which are generally used for downstream tasks, would not
    benefit from the disentanglement that was exhibited by sampled representations.~However, we showed that active variables, the relevant part of the
    representations, are quite similar in both representations.~Thus, we can expect mean
    representations to be as useful as sampled ones for downstream tasks.~We established that the passive variables of
    mean representations are near zero with low variance and seem to have an arbitrary high correlation with other variables.
    Thus, one may want to remove them, especially when feeding the mean representations into algorithms that are sensitive to
    near-zero or highly correlated features.

    \paragraph{Generalisation of our results to other types of VAEs}
    Because our paper explains the reason of the discrepancies between the mean and sampled representations observed by~\citet{Locatello2019a},
    we chose to remain as close as possible to their experimental protocol, and used the same models to obtain consistent results.
    However, the authors proved that unsupervised disentanglement learning was not possible without further inductive biases.
    Thus one should keep in mind that disentangled representations obtained from the models used in this paper can only be
    selected post-hoc based on disentanglement metrics, which may not always be practical for downstream task applications.
    Despite this, as described in Section 3, our explanation mainly relies on the polarised regime, which has been shown to
    occur in any well-behaved VAE as long as their prior and posterior distributions are Gaussian with diagonal covariances
    \citep{Dai2018,Rolinek2019,Dai2020,Bonheme2023a}.
    This makes our findings to readily applicable to identifiable VAEs which can directly provide
    disentangled representations~\citep{Khemakhem2020,Mita2021}.

    \paragraph{Other applications of this study} While our main focus was to explain the higher correlation observed in
    mean representations by \citet{Locatello2019a}, our new definitions of the polarised regime may also be useful
    to monitor the number of passive variables and prevent posterior collapse due to over-regularisation~\citep{Lucas2019b, Dai2020}.

    \paragraph{Future work} The surprising behaviour of annealed VAEs discussed in~\Appref{sec:annealed-vae}
    and the correlation of passive variables reported in~\Secref{sec:passive-variables} show that we could gain a deeper
    understanding of the representations learned by VAEs by studying their learning dynamics more in depth.
    While this was not the topic of this paper, we plan to address it in our future work.~Indeed, we believe that this
    could give more insight into how VAEs learn and how we can improve the learned representations.

    \vskip 0.6in

    \acks{We thank our action editor and reviewers for their detailed and constructive feedback, which helped us improve the quality of the present paper.}

    \newpage

    \appendix

    \section{Derivations and proofs}\label{sec:calc-app}

    \subsection{Details of the simplification of~\Eqref{eq:tc-gaussian-raw} to~\Eqref{eq:tc-gaussian}}\label{subsec:calc-kl-tc}
    \Eqref{eq:tc-gaussian-raw} has the following form
    \begin{equation*}
        TC(\rvr) = \klBigg{\N(\bar{\rvr}, \Cov[\rvr])}{\prod_{j=1}^d \N(\bar{\rvr}_j, \Cov[\rvr]_{jj})}.
    \end{equation*}
    We assume that all the distribution have a mean of zero. Thus,
    \begin{equation*}
        TC(\rvr) = \klBigg{\N(\vzero, \Cov[\rvr])}{\prod_{j=1}^d \N(0, \Cov[\rvr]_{jj})}.
    \end{equation*}
    Moreover, the second term of the KL divergence represents the case where $\rvr$ is composed of $j$ independent and
    normally distributed random variables. We can thus reexpress it as a multivariate Gaussian distribution with diagonal covariance,
    \begin{equation*}
        TC(\rvr) = \klBig{\,\N(\vzero, \Cov[\rvr])}{\N(\vzero, diag[\Var(\rvr)])\,}.
    \end{equation*}
    Now, let us define $\mSigma \triangleq \Cov[\rvr]$ and $\bar{\mSigma} \triangleq diag[\Var(\rvr)]$. Using the analytical
    solution of the KL divergence between two multivariate Gaussian distributions, we have:
    \begin{equation*}
        TC(\rvr) = \frac{1}{2} \left( \log \frac{\det(\bar{\mSigma})}{\det(\mSigma)} + \Tr(\bar{\mSigma}^{-1}\mSigma) - n \right).
    \end{equation*}
    Because $\mSigma$ and $\bar{\mSigma}$ have the same diagonal values,
    $\bar{\mSigma}^{-1}\mSigma = \mI$ and $ \Tr(\bar{\mSigma}^{-1}\mSigma) = n$. Thus,
    \begin{equation*}
        TC(\rvr) = \frac{1}{2} \left( \log \det(\bar{\mSigma}) - \log \det(\mSigma) \right).
    \end{equation*}
    As $\bar{\mSigma}$ is diagonal, this further simplifies to
    \begin{align*}
        TC(\rvr) &= \frac{1}{2} \bigg( \log \prod_{j=1}^{d}\bar{\mSigma}_{jj} - \log \det(\mSigma) \bigg),\\
        &= \frac{1}{2} \bigg( \sum_{j=1}^{d} \log \bar{\mSigma}_{jj} - \log \det(\mSigma) \bigg),\\
        &= \frac{1}{2} \bigg( \sum_{j=1}^{d} \log \Var(\rvr_j) - \log \det(\Cov[\rvr]) \bigg),
    \end{align*}
    as expected.

    \subsection{Proof of~\Thref{thm:polarised-sampled}}\label{subsec:proof-polarised-sampled}
    \begin{proof}
        Let $\rvz_j$ be the sampled representation variable at index $j$. There are three cases:
        \begin{enumerate}
            \item If $j \in \sV_p$, then, from statement (i) of~\Propref{prop:polarised-mean}, $\rvmu_j$ is almost constant with a value close to 0. Thus,
            $\rvz_j \approx \rvepsilon_j \rvsigma_j^{1/2}$. Using statement (i) of
            ~\Propref{prop:polarised-var}, we also know that $\rvsigma_j$ is almost constant with a value close to 1, thus $\rvsigma_j^{1/2} \approx 1$ and $\rvz_j \approx \rvepsilon_j$. It follows that $p(\rvz_j) \approx p(\rvepsilon_j)$, which proves statement (i).
            \item If $j \in \sV_a$, then, from statement (ii) of~\Propref{prop:polarised-var}, $\rvsigma_j$ is almost constant with a value close to 0. Thus, $\rvz_j \approx \rvmu_j$. It follows that $p(\rvz_j) \approx p(\rvmu_j)$, which proves statement (ii).
            \item If $j \in \sV_m$, then from statement (iii) of~\Defref{def:var-type} we know that $\rvz_j$ is composed of a subset of active components and a subset of passive components. Thus, $\rvz_j$ is distributed according to a mixture distribution. Using step (i) and (ii) of the proof, we know that $p(\rvz_j) \approx p(\rvepsilon_j)$ for passive variables and $p(\rvz_j) \approx p(\rvmu_j)$ for active variables.
            It follows that for mixed variables $p(\rvz_j) = c\; p(\rvepsilon_j) + (1 - c) \; p(\rvmu_j)$ where $0 < c < 1$. This concludes the proof.
        \end{enumerate}
    \end{proof}

    \subsection{Proof of~\Thref{thm:tc-pv}}\label{subsec:proof-tc-pv}
    \begin{proof}
        Let us consider a sampled representation $\rvz$ with $n$ latent variables having a covariance matrix $\Cov[\rvz] \in \R^{n \times n}$.
        Now, let us create a second sampled representation $\hat{\rvz}$ by concatenating the latent variables of $\rvz$ with $m$ passive variables.
        The resulting covariance matrix $\Cov[\hat{\rvz}]$ can be partitioned as
        \begin{equation*}
            \Cov[\hat{\rvz}] = \begin{bmatrix}
                                   \Cov[\rvz]   & \vzero_{n,m} \\
                                   \vzero_{m,n} & \mI_{m,m}
            \end{bmatrix}.
        \end{equation*}\\
        From this, we can immediately see that
        \begin{equation}
            \begin{split}
                \sum_{i=1}^{n+m}(\log\Cov[\hat{\rvz}]_{ii}) &= \sum_{i=1}^{n}(\log\Cov[\rvz]_{ii}) + \sum_{i=1}^{m}(\log\mI_{ii}) \\
                &= \sum_{i=1}^{n}(\log\Cov[\rvz]_{ii}).
            \end{split}\label{eq:sum-equiv}
        \end{equation}
        Moreover, as $\Cov[\rvz]$ is invertible, using Schur's identity~\citep{Brualdi1983}:
        \begin{equation*}
            \begin{split}
                \det(\Cov[\hat{\rvz}]) &= \det(\Cov[\rvz])\det(\mI - \vzero_{m,n}\Cov[\rvz]^{-1}\vzero_{n,m})\\
                &= \det(\Cov[\rvz])\det(\mI)\\
                &= \det(\Cov[\rvz]).
            \end{split}
        \end{equation*}
        Thus,
        \begin{equation}
            \log{\det(\Cov[\hat{\rvz}])} = \log{\det(\Cov[\rvz])}.\label{eq:det-equiv}
        \end{equation}
        Recall from \Eqref{eq:tc-gaussian} that
        \begin{equation}
            TC(\hat{\rvz}) = \frac{1}{2}\left(\sum_{i=1}^{n+m}(\log\Cov[\hat{\rvz}]_{ii}) - \log{\det(\Cov[\hat{\rvz}])} \right).\label{eq:tc-app}
        \end{equation}
        Using the result of \Eqref{eq:sum-equiv}, we can replace the first term of \Eqref{eq:tc-app}, so that
        \begin{equation}
            TC(\hat{\rvz}) = \frac{1}{2}\left(\sum_{i=1}^{n}(\log\Cov[\rvz]_{ii}) - \log{\det(\Cov[\hat{\rvz}])} \right).\label{eq:tc-sum-equiv}
        \end{equation}
        Finally, we can replace the second term of \Eqref{eq:tc-sum-equiv} using \Eqref{eq:det-equiv} to obtain
        \begin{equation*}
            \begin{split}
                TC(\hat{\rvz}) &= \frac{1}{2}\left(\sum_{i=1}^{n}(\log\Cov[\rvz]_{ii}) - \log{\det(\Cov[\rvz])} \right),\\
                &= TC(\rvz),
            \end{split}
        \end{equation*}
        as required.
    \end{proof}

    \subsection{Proof of~\Thref{thm:mi-pv}}\label{subsec:proof-mi-pv}
    \begin{proof}
        Let us consider a sampled representation $\rvz \in \R^2$ composed of two active variables $\rvz_1$ and $\rvz_2$ such that
        $MI(\rvz_1, \rvz_2) = c$ with $c > 0$. As $MI$ is symmetric, we have $MI_{avg}(\rvz) = \frac{1}{2} 2c = c$.\\
        Now, let us consider a sampled representation $\hat{\rvz} \in \R^{n}$ composed of the two actives variables of $\rvz$ and
        $n-2$ additional passive variables $\{\rvz_j\}_{j=3}^n$.\\
        Because passive variables do not contain any information about the input, the mutual information between an active variable $i$
        and a passive variable $j$ will be zero (i.e., $MI(\rvz_i, \rvz_j) = 0$).
        Passive variables are also independent and normally distributed, hence the mutual information between two different passive
        variables $i$ and $j$ will also be zero.
        We thus have
        \begin{equation*}
            \begin{split}
                MI_{avg}(\hat{\rvz}) &= \frac{1}{n^2 - n} \left(\sum_{i \in \sV_a}\sum_{\substack{j \in \sV_a\\i \neq j}} MI(\rvz_i, \rvz_j) + 2\sum_{i \in \sV_a}\sum_{j \in \sV_p} MI(\rvz_i, \rvz_j) + \sum_{i \in \sV_p}\sum_{\substack{j \in \sV_p\\i \neq j}} MI(\rvz_i, \rvz_j)\right),\\
                &= \frac{1}{n^2 - n} 2c < MI_{avg}(\rvz),
            \end{split}
        \end{equation*}
        as required.
    \end{proof}

    \section{The curious case of Annealed VAE}\label{sec:annealed-vae}
\begin{figure}[ht!]
        \centering
        \begin{subfigure}{.32\textwidth}
            \centering
            \includegraphics[width=\linewidth,keepaspectratio]{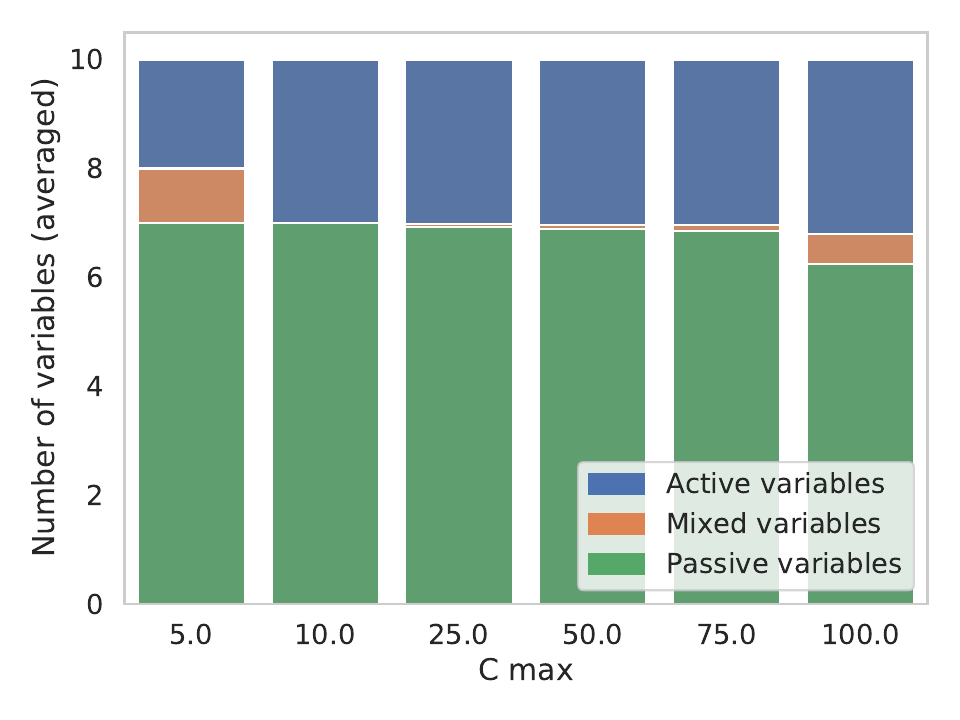}
            \caption{DSprites}
        \end{subfigure}
        \begin{subfigure}{.32\textwidth}
            \centering
            \includegraphics[width=\linewidth,keepaspectratio]{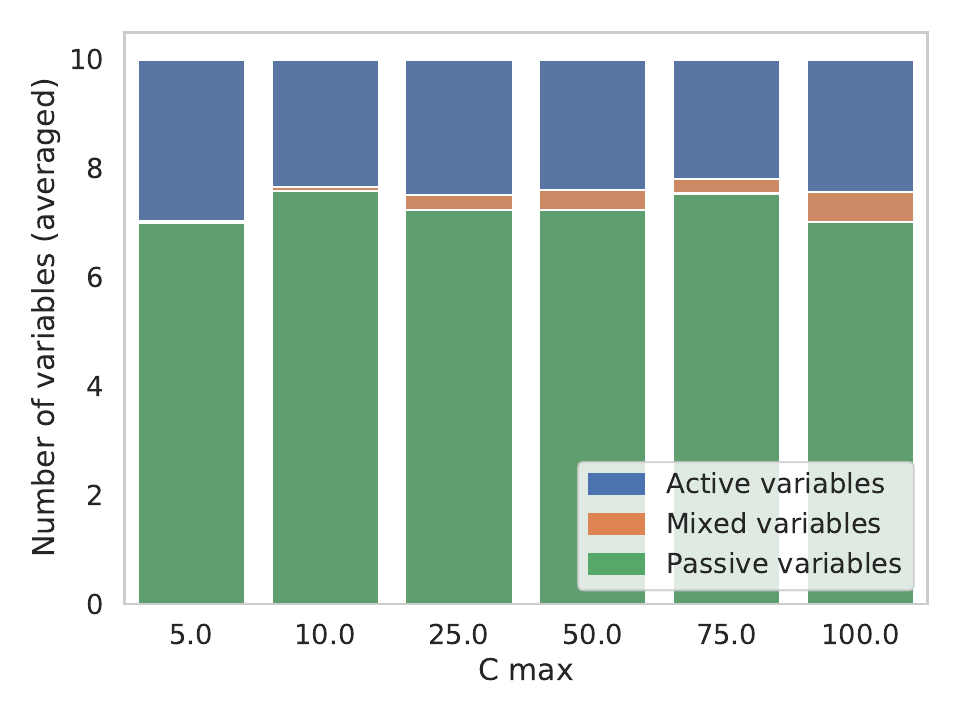}
            \caption{Scream dSprites}
        \end{subfigure}
        \begin{subfigure}{.32\textwidth}
            \centering
            \includegraphics[width=\linewidth,keepaspectratio]{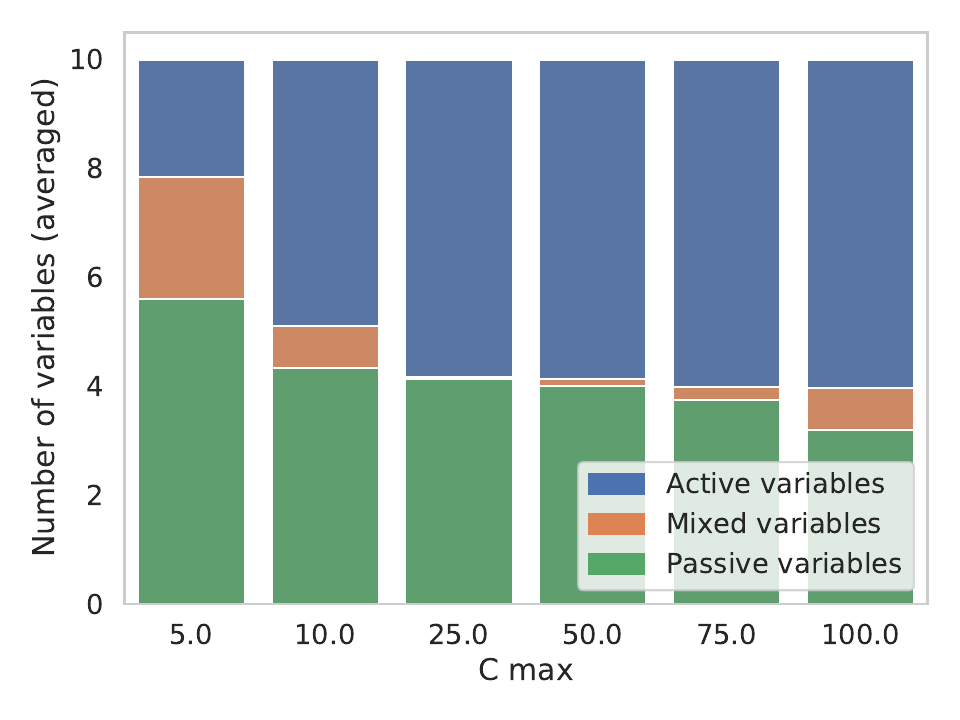}
            \caption{Cars3D}
        \end{subfigure}
        \caption{Number of passive, mixed and active variables of annealed VAE trained on dSprites, scream dSprites and cars3D
        with decreased regularisation strength. The results are averaged over 50 runs for each regularisation value.}
        \label{fig:passive-var-annealed}
    \end{figure}

    In opposition to the other models that we studied in this paper, annealed VAE surprisingly exhibits a high number of passive variables regardless of the
    regularisation strength on most datasets, which can be seen in \Figref{fig:passive-var-annealed}.
    Note that in contrast to the remaining architectures that we study, a higher value of the hyperparameter \textrm{C} means that the regularisation strength decreases, whereas higher $\beta$ in $\beta$-VAEs implies stronger regularisation.
    As~\citet{Burgess2018} originally argued that a higher channel capacity, \textrm{C}, should help the model to learn more latent factors as the training progresses,
    one would assume that the number of active variables should increase with a higher value of \textrm{C}, but it is generally not the case in \Figref{fig:passive-var-annealed}.

\begin{figure}[ht!]
        \centering
        \begin{subfigure}{.4\textwidth}
            \centering
            \includegraphics[width=\linewidth,keepaspectratio]{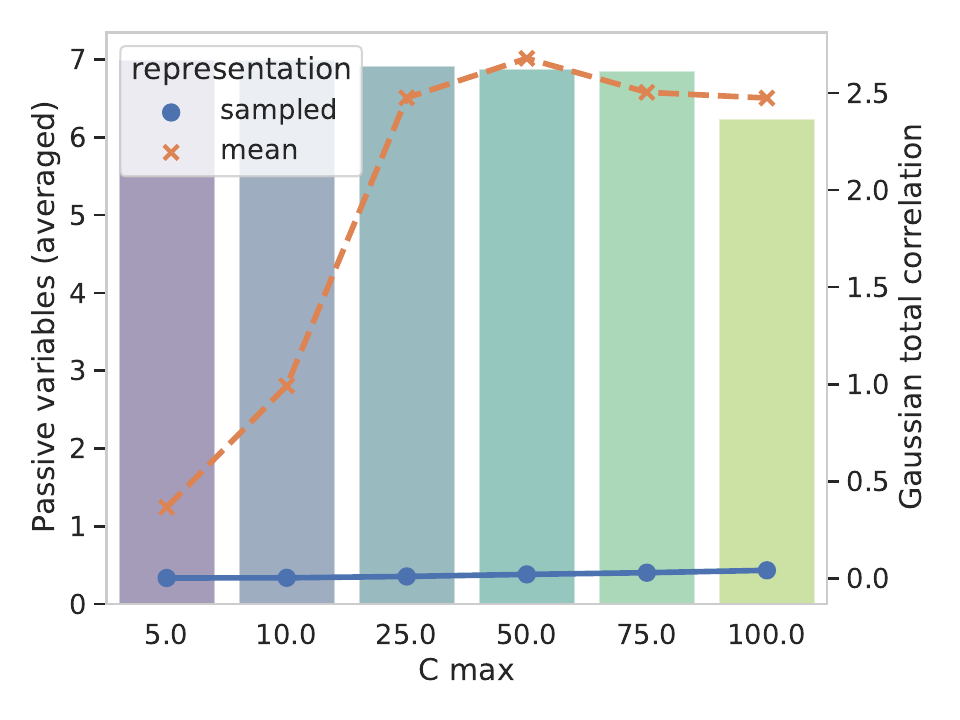}
            \caption{}
            \label{subfig:tc-dsprites}
        \end{subfigure}
        \begin{subfigure}{.4\textwidth}
            \centering
            \includegraphics[width=\linewidth,keepaspectratio]{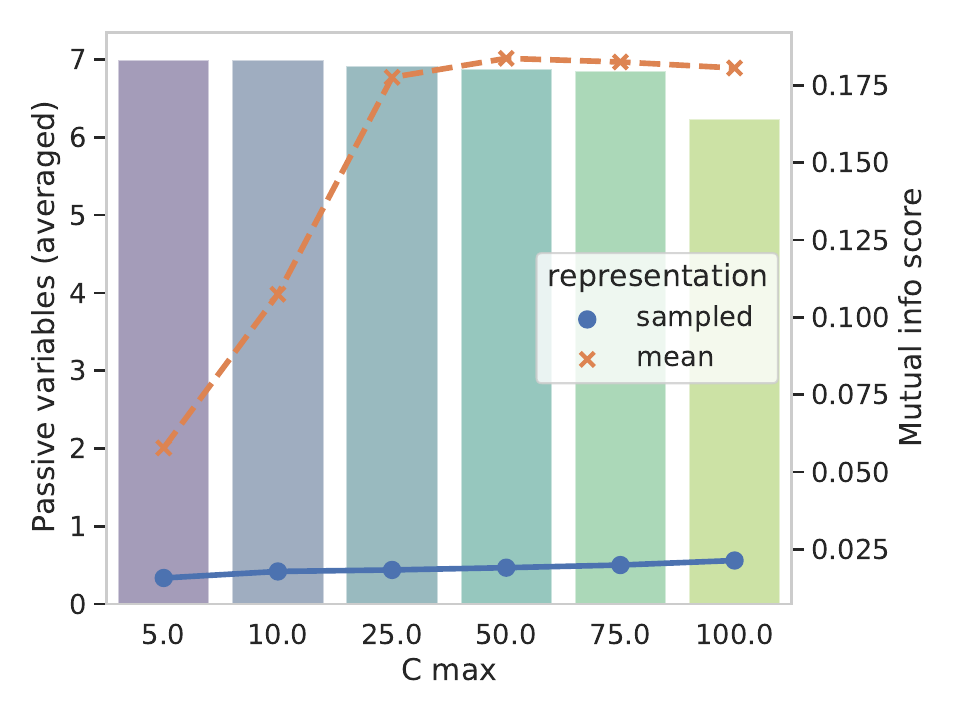}
            \caption{}
            \label{subfig:mi-dsprites}
        \end{subfigure}
        \caption{Comparison of the total correlation and averaged mutual information with the number of passive variables of mean and sampled representations of annealed VAE trained on dSprites.
        Figure (a) is the total correlation and Figure (b) the averaged mutual information. The lines indicate the metric scores of the two representations, and the bars the average number of passive variables.}
        \label{fig:tc-pv-evol-annealed}
    \end{figure}

\begin{figure}[ht!]
        \centering
        \begin{subfigure}{.4\textwidth}
            \centering
            \includegraphics[width=\linewidth,keepaspectratio]{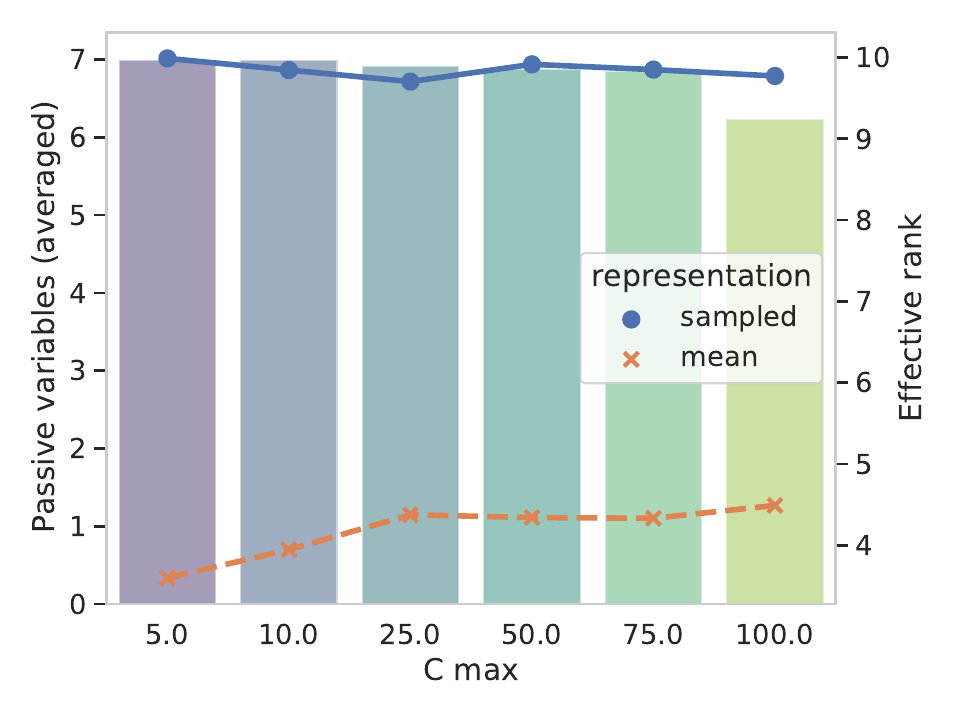}
            \caption{}
            \label{subfig:er-dsprites}
        \end{subfigure}
        \begin{subfigure}{.4\textwidth}
            \centering
            \includegraphics[width=\linewidth,keepaspectratio]{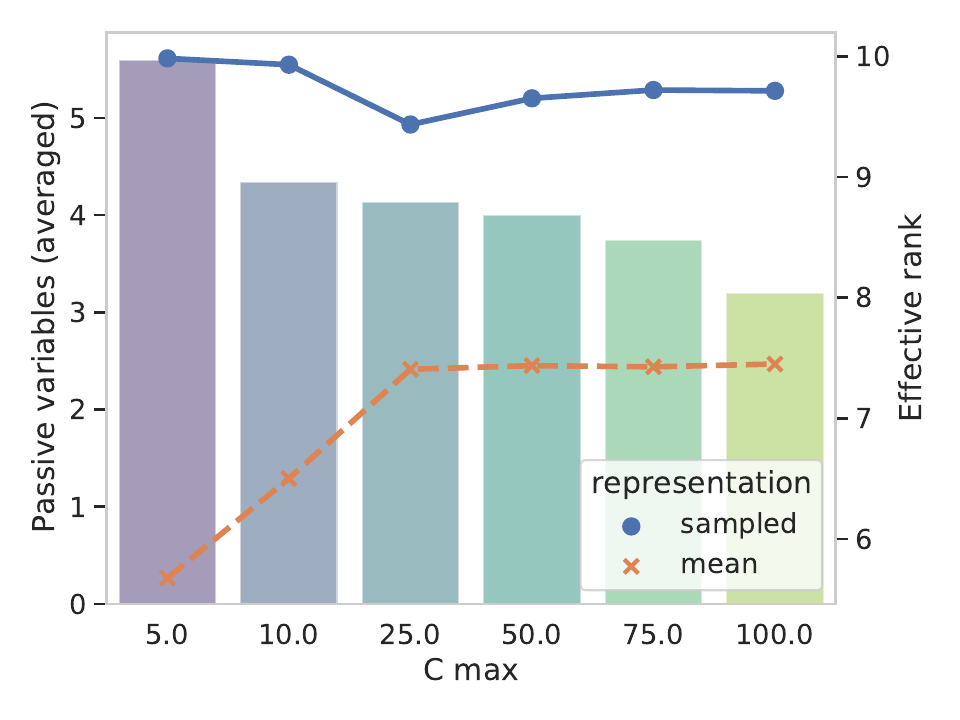}
            \caption{}
            \label{subfig:er-cars3d}
        \end{subfigure}
        \caption{Comparison of the effective rank with the number of passive variables of mean and sampled representations of annealed VAE trained on dSprites and Cars3D.
        The lines indicate the metric scores of the two representations, and the bars the average number of passive variables.}
        \label{fig:er-pv-evol-annealed}
    \end{figure}

\begin{figure}[ht!]
        \centering
        \begin{subfigure}{.4\textwidth}
            \centering
            \includegraphics[width=\linewidth]{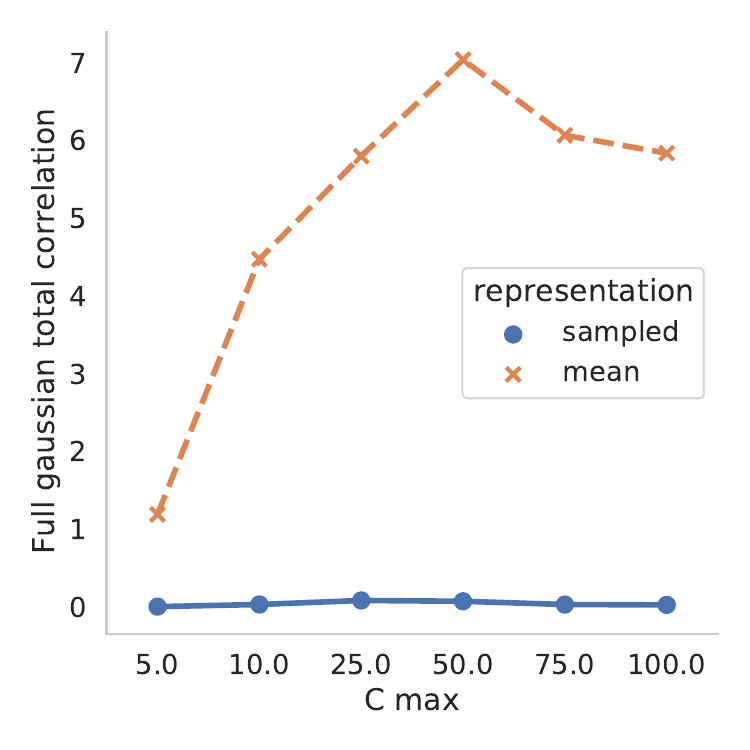}
            \caption{}
        \end{subfigure}\begin{subfigure}{.4\textwidth}
                           \centering
                           \includegraphics[width=\linewidth]{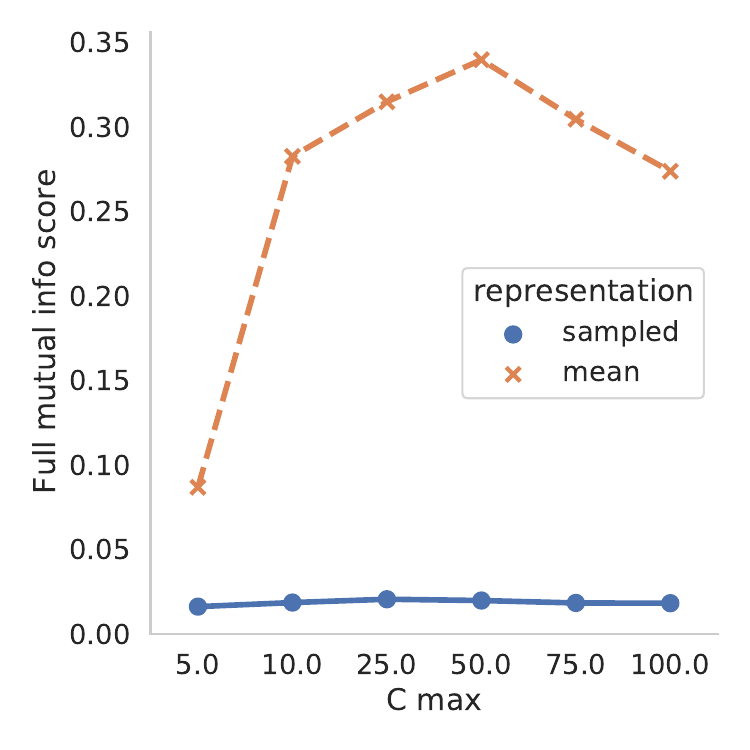}
                           \caption{}
        \end{subfigure}\\
        \begin{subfigure}{.4\textwidth}
            \centering
            \includegraphics[width=\linewidth]{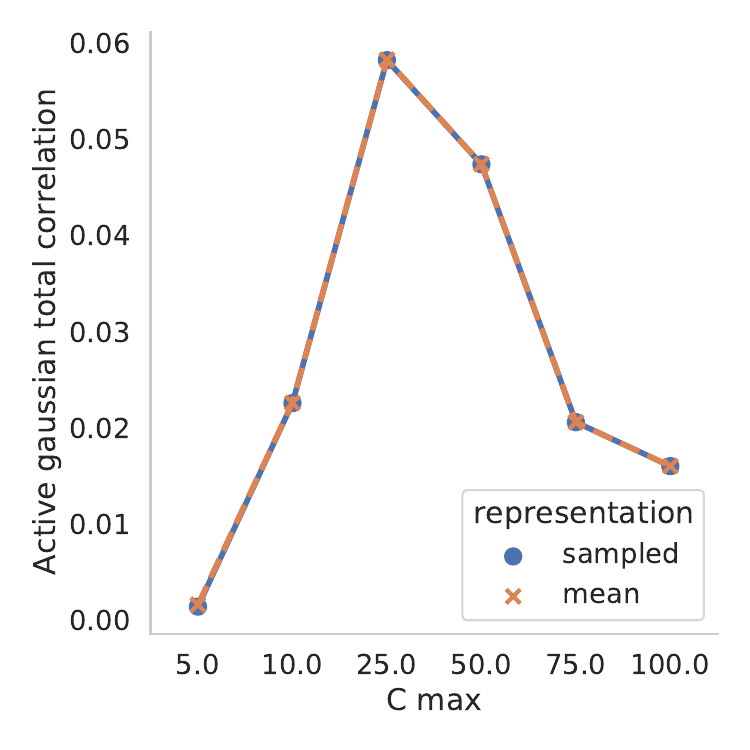}
            \caption{}
        \end{subfigure}
        \begin{subfigure}{.4\textwidth}
            \centering
            \includegraphics[width=\linewidth]{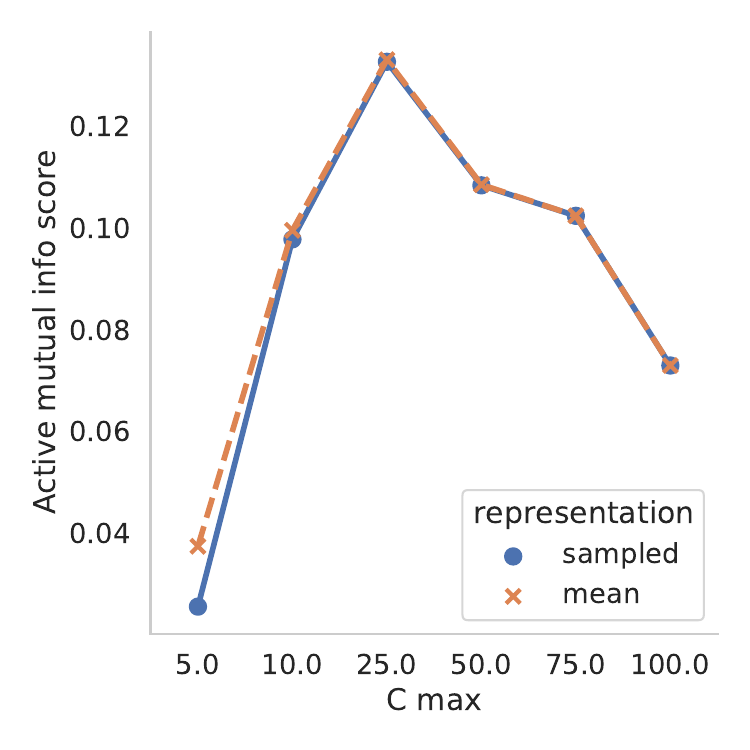}
            \caption{}
        \end{subfigure}
        \caption{Comparison of the total correlation and averaged mutual information scores of the mean representation of annealed VAE trained on scream dSprites.
        Figures (a) and (b) are the results of the total correlation and averaged mutual information score using the full representation, and figures (c) and (d) are
        the results using active variables only.}
        \label{fig:comb-tc-annealed-scream-dsprites}
    \end{figure}

    Given the near constant number of passive variables observed across all the bars in \Figref{fig:passive-var-annealed}, one could expect similarly constant TC and
    averaged MI scores.~However, we can see in \Twofigref{fig:tc-pv-evol-annealed}{fig:tc-pv-evol-annealed2} that
    the TC and averaged MI generally decrease with higher regularisation strengths.~Moreover, \Figref{fig:comb-tc-annealed-scream-dsprites} shows that removing the passive
    variables effectively reduces the overall TC and averaged MI scores, suggesting that a higher channel capacity, \textrm{C}, (i.e., a lower regularisation strength at the end of training) may encourage
    the passive variables of mean representations to be more correlated.
    This is further confirmed in~\Figref{fig:er-pv-evol-annealed}, where
    we can see that the effective rank obtained with a higher channel capacity and less passive variables is close to
    the one obtained with a lower channel capacity and more passive variables.
    For example, in cars3D, the effective rank of the mean representation for a channel capacity of 25 and 4 passive variables
    is the same as the one obtained for a channel capacity of 100 and 3 passive variables.
    In conclusion, despite their near-constant number of passive variables, annealed VAE's results are consistent with our other
    findings: the passive variables of the mean representations are still responsible for the higher TC and averaged MI
    scores, but their correlation seems to increase with the channel capacity, \textrm{C}.

    \begin{figure}[ht!]
        \centering
        \begin{subfigure}{.4\textwidth}
            \centering
            \includegraphics[width=\linewidth,keepaspectratio]{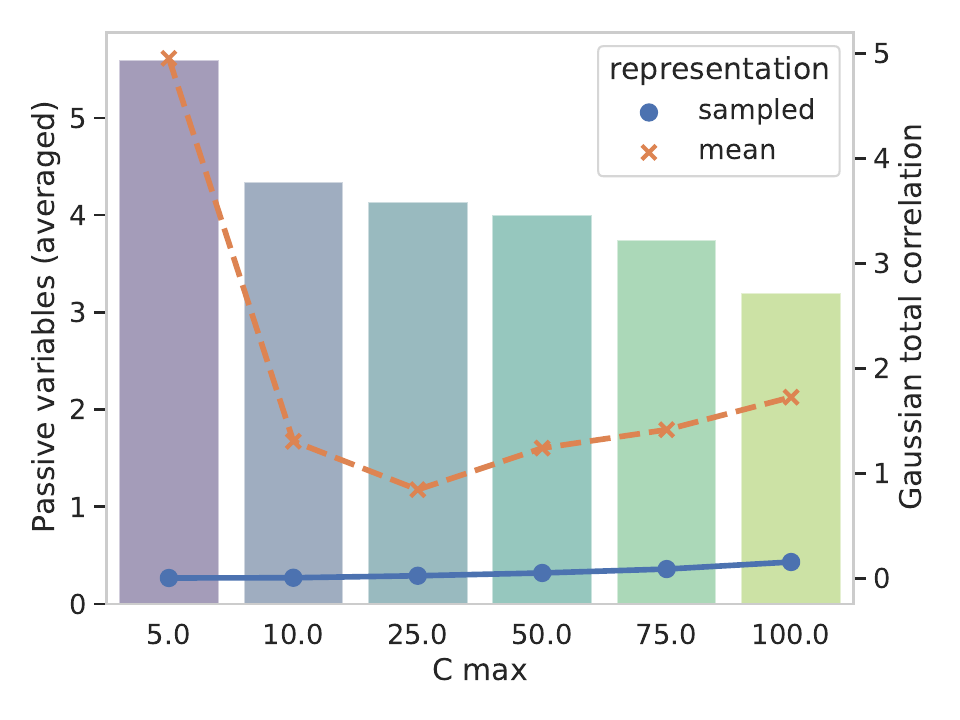}
            \caption{}
            \label{subfig:tc-cars3d}
        \end{subfigure}
        \begin{subfigure}{.4\textwidth}
            \centering
            \includegraphics[width=\linewidth,keepaspectratio]{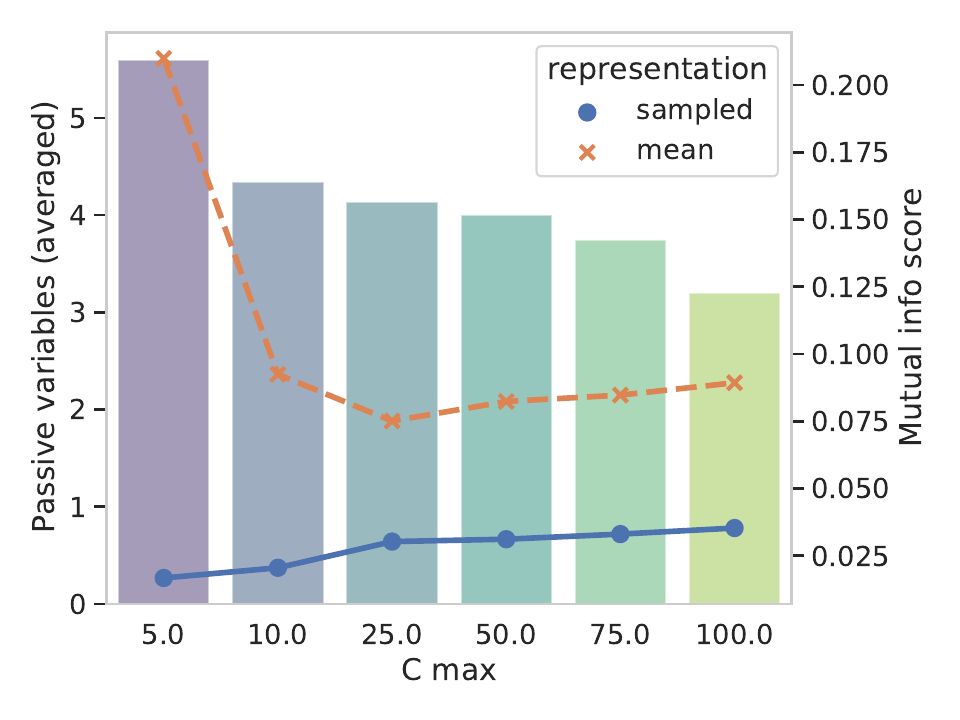}
            \caption{}
            \label{subfig:mi-cars3d}
        \end{subfigure}
        \caption{Comparison of the total correlation and averaged mutual information with the number of passive variables of mean and sampled representations of annealed VAE trained on Cars3D.
        Figure (a) is the total correlation and Figure (b) the averaged mutual information. The lines indicate the metric scores of the two representations, and the bars the average number of passive variables.}
        \label{fig:tc-pv-evol-annealed2}
    \end{figure}

    \clearpage
    \bibliography{bibliography}

\end{document}